\documentclass[sn-basic]{sn-jnl}

\usepackage{graphicx}%
\usepackage{multirow}%
\usepackage{amsmath,amssymb,amsfonts}%
\usepackage{amsthm}%
\usepackage{booktabs}%
\usepackage{mathrsfs}%
\usepackage[title]{appendix}%
\usepackage{xcolor}%
\usepackage{textcomp}%
\usepackage{manyfoot}%
\usepackage{algorithm}%
\usepackage{algorithmicx}%
\usepackage{algpseudocode}%
\usepackage{listings}%
\usepackage{orcidlink}

\newcommand{\dashrule}[1][black]{%
  \color{#1}\rule[\dimexpr.5ex-.2pt]{4pt}{.4pt}\xleaders\hbox{\rule{4pt}{0pt}\rule[\dimexpr.5ex-.2pt]{4pt}{.4pt}}\hfill\kern0pt%
}

\newcommand{\parafango}[1]{\textbf{#1.$\ $}}

\theoremstyle{thmstyleone}%
\theoremstyle{thmstyletwo}%

\theoremstyle{thmstylethree}%

\begin{document}

\title[The KANDY Benchmark]{The KANDY Benchmark: Incremental Neuro-Symbolic Learning and Reasoning with Kandinsky Patterns}


\author*[1,2]{\fnm{Luca Salvatore} \sur{Lorello} \textsuperscript{\orcidlink{0009-0006-3803-8268}}}\email{luca.lorello@phd.unipi.it}

\author[2]{\fnm{Marco} \sur{Lippi} \textsuperscript{\orcidlink{0000-0002-9663-1071}}}\email{marco.lippi@unimore.it}

\author[3]{\fnm{Stefano} \sur{Melacci} \textsuperscript{\orcidlink{0000-0002-0415-0888}}}\email{mela@diism.unisi.it}

\affil[1]{\orgdiv{Department of Computer Science}, \orgname{University of Pisa}, \orgaddress{\street{Largo B.~Pontecorvo 3}, \city{Pisa}, \postcode{56127}, \country{Italy}}}

\affil[2]{\orgdiv{Department of Sciences and Methods for Engineering}, \orgname{University of Modena and Reggio Emilia}, \orgaddress{\street{via Amendola 2}, \city{Reggio Emilia}, \postcode{42122}, \country{Italy}}}

\affil[3]{\orgdiv{Department of Information Engineering and Mathematics}, \orgname{University of Siena}, \orgaddress{\street{via Roma 56}, \city{Siena}, \postcode{53100}, \country{Italy}}}


\abstract{Artificial intelligence is continuously seeking novel challenges and benchmarks to effectively measure performance and to advance the state-of-the-art. In this paper we introduce KANDY, a benchmarking framework that can be used to generate a variety of learning and reasoning tasks inspired by Kandinsky patterns. By creating curricula of binary classification tasks with increasing complexity and with sparse supervisions, KANDY can be used to implement benchmarks for continual and semi-supervised learning, with a specific focus on symbol compositionality. Classification rules are also provided in the ground truth to enable analysis of interpretable solutions.
Together with the benchmark generation pipeline, we release two curricula, an easier and a harder one, that we propose as new challenges for the research community.
With a thorough experimental evaluation, we show how both state-of-the-art neural models and purely symbolic approaches struggle with solving most of the tasks, thus calling for the application of advanced neuro-symbolic methods trained over time.}




\maketitle

\section{Introduction}\label{sec:intro}

Properly evaluating the performance of Artificial Intelligence (AI) systems is crucial to measure progress in the area as well as to identify limitations even in state-of-the-art models.
In particular, in the last years there has been a growing interest in developing novel benchmarks that can assess the capability of AI systems to perform reasoning tasks that require skills beyond plain pattern recognition~\citep{chollet2019measure}.
Such problem has become prominent in the neuro-symbolic (NeSy) AI community, for specific reasons. First of all, NeSy approaches aim to combine the best of two different worlds, namely neural models to handle and process perceptual stimuli, and symbolic methods to deal with knowledge representation and reasoning. Therefore, to prove the advantages of these hybrid models, dedicated benchmarks are often required, that should highlight the limitations of the two families of approaches, taken separately~\citep{ott2023think}.\footnote{To remark the importance of designing novel benchmarks, in the context of the TAILOR network of excellence (\url{https://tailor-network.eu/}) a workshop on benchmark development was organized within the NeSy 2023 conference: \url{https://sailab.diism.unisi.it/tailor-wp4-workshop-at-nesy/}}
Thanks to an explicit representation of knowledge (typically through logic), NeSy approaches inherently favour interpretability, which is widely recognized to be a crucial feature to achieve trustworthy AI~\citep{kaur2022trustworthy}.
As an additional challenge, many NeSy frameworks are currently trying to incorporate continual and semi-supervised settings in their learning paradigms~\citep{DBLP:conf/icml/MarconatoBFCPT23,liu2023weakly,yin2022visual}: an issue that also needs benchmarks with specific characteristics.

In this paper, we propose a novel benchmark to implement agents which can incrementally build hierarchical representations that exploit compositionality. We aim to target problems that are typically simple for humans but often very difficult for machines, and especially for neural networks: examples of such tasks are given by Bongard problems~\citep{youssef2022towards}, Michalski's trains~\citep{helff2023v,michalski1980pattern}, and Kandinsky patterns~\citep{holzinger2019kandinsky}. We conjecture that this kind of tasks are better addressed by machines that acquire the needed skills progressively in a continual setting, guided by a teacher that opportunely gives sporadic supervisions to incrementally learn novel concepts by composing the ones acquired earlier.\footnote{Besides traditional studies in curriculum learning~\citep{wang2021survey}, the use of a teaching paradigm provided by a machine has been recently successfully explored also in human learning~\citep{ai2023explanatory}.} To implement our benchmark, we choose Kandinsky-like problems, as they are becoming very popular in the NeSy community~\citep{shindo2023alpha} and offer a controlled  simple environment, which allows to design very complex and challenging scenarios.

We release two versions of the benchmark: a first one that contains a collection of ``easy'' tasks, and a second, ``hard'' one with much more challenging problems. In both cases, we organize tasks according to a curriculum-like paradigm, and we implement a strategy to control the amount of supervision, ranging from fully supervised data to sporadically provided labels. Furthermore, our framework is intended to be both used as is and extended by users to generate their own benchmarks: to enable this feature, we provide a tool that easily allows task generation by exploiting a set of pre-defined atomic objects, with a few customizable properties, and compositional primitives that make it possible to create concepts with increasing complexity.

To summarize, we hereby list the main contributions of this work, remarking how several communities could actually benefit from the novel benchmark we propose. 
%
\begin{itemize}
\item We release a benchmark that combines visual scene understanding, in a context that exhibits the binding problem~\citep{treisman1998feature}, with an inductive reasoning component. Both symbolic and sub-symbolic approaches can be separately evaluated on the benchmark: by showing that each of the two families of methods has specific limitations, we advocate the use of NeSy frameworks.
\item The benchmark allows to assess the explainability of the applied methodologies: in fact, all the classification problems can be explained in terms of logic rules, and by releasing the ground truth we allow the diagnosis of learned rules.
\item We focus on a continual learning setting, with a teacher model that densely or sparsely provides supervisions, thus opening to the evaluation of semi-supervised and continual learning approaches in the NeSy community.
\item We implement object compositionality, building on a collection of simple objects and basic primitives, extending perception to recursive structures. 
\item We release two curricula implementing the benchmark 
where tasks are progressively ordered by difficulty, so that more complex concepts and tasks are built on top of the ones learned earlier in the curriculum. 
\item We allow end-users to easily extend or fully customize the benchmark, by creating their own tasks and curricula using a generation tool.
\end{itemize}

The paper is structured as follows. In Section~\ref{sec:related} we discuss related works in the context of abstract reasoning benchmarks. Section~\ref{sec:benchmark} illustrates our proposal, its basic blocks and compositional primitives,  
and the publicly released benchmarks. 
Section~\ref{sec:experiments} reports experimental results obtained with both symbolic and sub-symbolic baselines, to emphasize the difficulties of the benchmark. Finally, Section~\ref{sec:conclusions} concludes the paper.



\section{Related works}\label{sec:related}

Visual reasoning tasks involving abstract shapes and objects, or simplified artificial scenes, have recently received a lot of attention, as they allow to create controlled environments where perceptual and reasoning skills can be easily tested and compared. 

\parafango{Kandinsky patterns}
Kandinsky patterns~\citep{muller2021kandinsky} are simple, procedurally generable and human understandable figures, consisting of elementary shapes arranged according to a certain criterion, which can be organized into data sets for the assessment of machine perception.
The KANDINSKYPatterns~\citep{holzinger2021kandinskypatterns} benchmark is a set of three challenges relying on the identification of spatial and Gestalt concepts, posed as binary classification tasks, where positives can be described by a complex rule, while negatives are perceptually very similar, but do not satisfy the rule.
The Elementary Concept Reasoning~\citep{stammer2022interactive} data set simplifies Kandinsky patterns to the extreme, images contain a single elementary concept, characterized by three properties: shape, size and (noisy) color, with the aim to assess generalizability and interactive learning of new concepts. 
%
Our benchmark is based on Kandinsky-like binary problems, but with a strong emphasis on compositionality, incremental learning, and sparse supervision. 

\parafango{Bongard problems}
The Bongard problems~\citep{bongard1970pattern} are a classic cognitive psychology test, in which humans are presented with few images coming from two sets. One of these sets (the positive set) is characterized by an arbitrarily complex property (e.g., ``all figures have lines of constant length'', or ``there are at least two intersecting circles''), while the other (the negative set) contains only images violating it.
Participants are asked to infer the hidden property characterizing the positive set, and describe it with a natural language statement, by observing a limited number of samples (e.g., six positive images and six negative ones).
Being notoriously complex for purely subsymbolic approaches~\citep{yun2020deeper}, Bongard problems have been recently proposed in a variety of different settings and scenarios.
Bongard-LOGO~\citep{nie2020bongard} is a synthetic benchmark directly inspired by the original Bongard problems, whereas
Bongard-HOI~\citep{jiang2022bongard} is a natural image benchmark: both are framed as a few-shot binary classification task. 
Bongard-Tool~\citep{jiang2023bongard} is a similar natural image extension of Bongard-LOGO, which requires stronger semantic knowledge than Bongard-HOI, in the form of few-shot concept induction.
Our benchmark is indeed based on discriminating images exploiting possibly sparse supervisions, but it is mostly focused on providing conditions for evaluating NeSy learning in an incremental setting, emphasizing compositionality of the progressively presented patterns. Although we relax the extreme few-shot learning characteristics of these benchmarks, we fully preserve the inductive nature of the original Bongard problems, which is lost in these purely discriminative benchmarks.

\parafango{Raven matrices}
A large body of work is dedicated to benchmarks based on Raven's progressive matrices~\citep{raven1938raven}, which are a non verbal psychometric test for general human intelligence and abstract reasoning, designed to be knowledge-agnostic and to be administered to the widest population possible, regardless of age, cultural background and mental conditions.
Participants are asked to solve a series of multiple choice questions, by selecting the correct pattern to complete a matrix (usually $3\times3$, missing the bottom right element).
Questions are presented with increasing difficulty and require counting, reasoning by analogy, pattern completion, visual operations (e.g., subtracting two figures), etc.
In the design of KANDY, we borrow the idea of abstract tasks, requiring the use of complex ``reasoning primitives'', presented in an increasingly harder fashion. We however relax the requirement of one-shot analogical reasoning, in favor of a curriculum-like organization where learned skills are reused in future tasks.
Some benchmarks have been inspired by Raven's progressive matrices. 
PGM~\citep{barrett2018measuring} is a benchmark 
which procedurally generates samples based on a symbolic representation of each problem.
%
RAVEN~\citep{zhang2019raven} is a similar benchmark, which instantiates far more diversified abstract reasoning rules than PGM, providing both a diagnostic symbolic structure 
and human baselines.
I-RAVEN~\cite{hu2021stratified} and RAVEN-Fair~\citep{benny2021scale} are more rigorous variants of RAVEN, which solve serious statistical biases in the original dataset.
Unicode Analogies~\citep{spratley2023unicode} is a RAVEN-like benchmark, which replaces simple features with unicode characters, introducing the need of Gestalt perception, and, to some extent, Bongard-like contrastive reasoning, to blur the line between object and feature, forcing models to embed contextual information both for perception and reasoning.
In a similar context, V-prom~\citep{teney2020v} extends Raven's progressive matrices to natural images.
%
 %
%
%
 %
%
%
Most benchmarks based on psychometric tests for humans suffer from a major flaw: as they are human-solvable and based on a limited set of rules, it is possible to inject knowledge about the benchmark's structure to achieve extremely high performance. For instance, the neuro-vector-symbolic architecture~\citep{hersche2023neuro} solves RAVEN and I-RAVEN with almost perfect performance, but implements the underlying problem structures as hard-coded operators on hyper-dimensional vectors.

\parafango{Other Benchmarks}
The literature includes other popular benchmarks in the context of abstract visual reasoning, from which KANDY borrows some ideas.
CLEVR~\cite{johnson2017clevr} is a large diagnostic dataset for visual question answering, with the explicit goal of overcoming the limitation of existing performance-based benchmark, which fail to pinpoint the reasons why a model succeeds or fails in answering a specific question.
Images in CLEVR are simple 3D scenes, rendered from a scene graph consisting of objects annotated with position 
and attributes
, and spatial relations with non-trivial semantics. 
Questions belong to parametric sets defined by a functional program (i.e., a graph of operations required to retrieve the correct answer) and one or more natural language templates associated with it.
Like CLEVR, KANDY contains diagnostic data in the form of symbolic representations for each sample, and ground truth rules for each task, allowing for deeper inspection of models.
Michalski's trains~\citep{larson1977inductive} are an inductive logic programming toy problem, in which trains are labeled with their traveling direction (``going east'' or ``going west''). Each train is composed by a certain number of wagons, characterized by a set of predicates (e.g., open/closed car, containing triangles, three sets of wheels, etc.) and the underlying assumption is that the label only depends on the properties of each train. The goal of the task is to determine the set of rules describing every train going in the two directions. Like Michalski's trains, KANDY is a collection of tasks where a binary decision rule is induced entirely from perceptual stimuli.
V-Lol~\cite{helff2023v} is a CLEVR-like dataset extending Michalski's trains to perceptual reasoning, with the objective of jointly evaluating exact logical reasoning, noisy scene understanding and abstract generalization.
The ARC~\citep{chollet2019measure} benchmark evaluates fluid intelligence both in humans and machine learning methods, without the requirement of language, visual or real-world common sense, or real-world objects knowledge.
Both inputs and outputs are discrete-color pixel grids and each sample corresponds to a different task.
Models are presented with three input/output pairs and are tasked to draw the output corresponding to a new input.

\section{The KANDY benchmark}\label{sec:benchmark}

\begin{figure}
\includegraphics[width=1.0\textwidth]{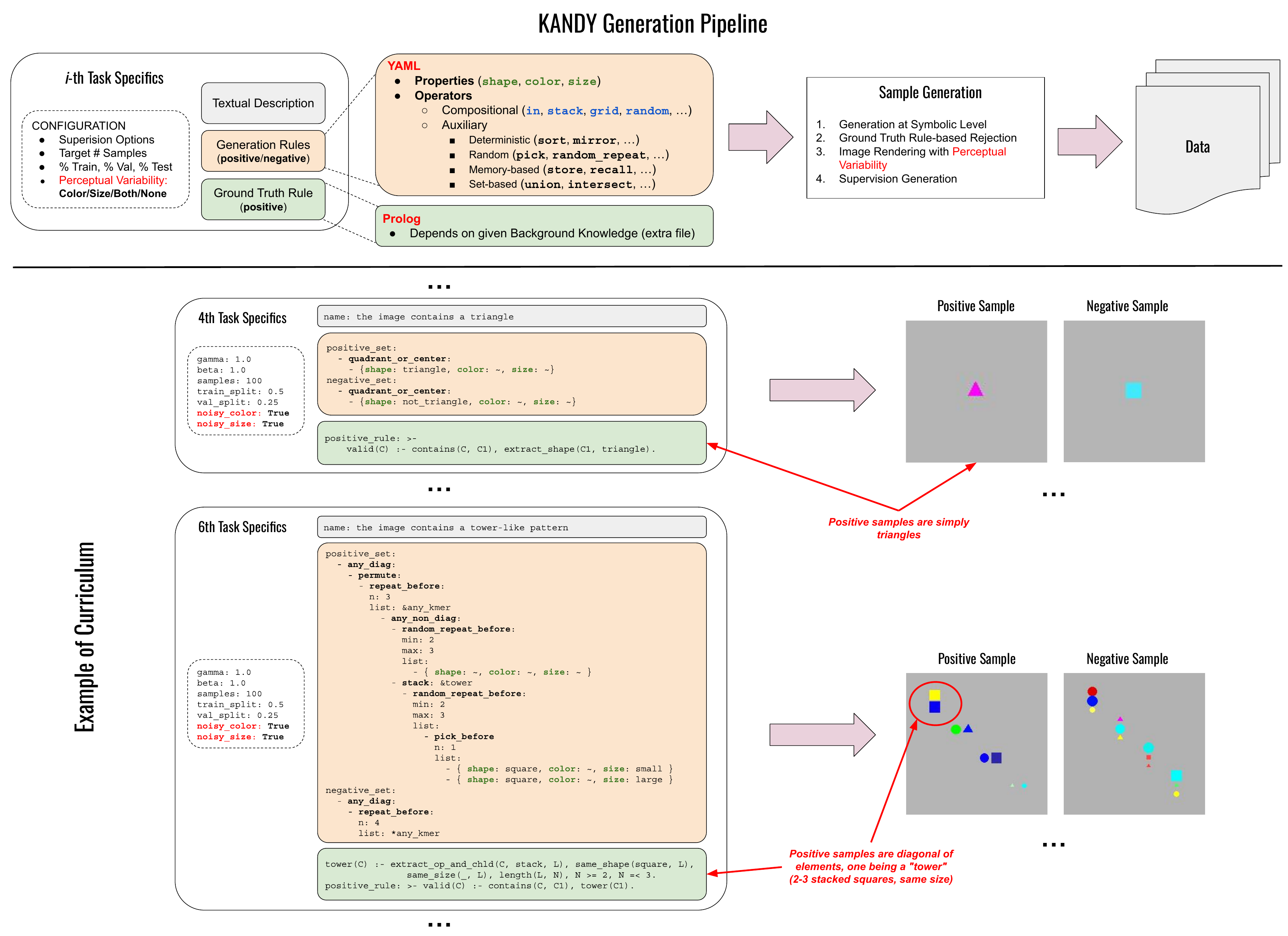}
\vskip -1mm
\caption{\textit{Top:} Overview of the KANDY generation pipeline: the user provides task specifics and data is generated. Positive and negative sets are defined via symbolic representations and rendered into synthetic images. The ground truth rule is a propositional clause that can be used to reject samples, and thus explains the task. \textit{Bottom:} Example of two tasks from a curriculum generated with KANDY. \label{fig:example-kandy}}
\end{figure}

KANDY is a benchmarking framework that users can exploit to generate curricula composed of sequences of tasks, where each task is a binary classification problem on image data, whose specifics are fully customizable. Before diving into the details of KANDY, we showcase its main possible usages.
\begin{enumerate}
    \item \textbf{Offline vs. continual learning.} Data generated by KANDY can be used to compare classical ``offline'' batch-mode learning, where the whole training data is immediately available to the learning agent, with continual/lifelong learning, where data is made progressively available to the learner, with or without repetitions.
    \item \textbf{Independent tasks vs. multi-task.} Tasks within a curriculum can be considered independent one from the other (thus training independent classifiers), or they can be jointly addressed in a multi-task setting (thus training a single classifier).
    \item \textbf{Curriculum vs. random order.} Tasks (and samples) can be shown to the learning agent in the order defined by the user (curriculum) or in a random order. In the former case, the agent is expected to progressively acquire skills to solve tasks of increasing complexity.
    \item \textbf{Supervised vs. semi-supervised.} Tasks can be fully supervised, as well as sparsely supervised with a customizable criterion, raising the additional challenge of dealing with a semi-supervised data set.
    \item \textbf{Interpretability vs. explainability.} Tasks and samples are paired with symbolic ground truth, allowing to compare interpretable-by-design systems and post-hoc explanations.
\end{enumerate}
In the following we describe the main building blocks of the framework, supported by Figure~\ref{fig:example-kandy}, and two curricula that we release as open challenges. 
Further details are provided in the supplementary material, while several usage-oriented details are  reported in the KANDY code repository, where we publicly share the data generator, our curricula, and code for running experiments with neural and symbolic baselines: \url{https://github.com/continual-nesy/KANDYBenchmark}.

\subsection{Data generation: basic blocks and primitives}\label{sec:generation}



Each task-related binary classification problem of a curriculum is defined by positive and negative sets of {\it generation rules} (YAML format), and optionally paired with a propositional clause that acts as a {\it ground-truth rule} (Prolog-interpretable $\texttt{valid/1}$ predicate, built on some background knowledge), which explains the task (Figure~\ref{fig:example-kandy}, top). 
KANDY yields synthetic images containing compositions of simple geometric shapes with several configurable {\it properties}: color, shape, and size. 
Images are generated accordingly to the aforementioned rules, that are described by means of different classes of {\it operators}. As a result, samples are associated with a symbolic tree, where leaves are atomic shapes and every other node is a composition of children. 
Images are iteratively generated by a sampling procedure, being possibly rejected when not coherent with the ground-truth rule, if provided.

\parafango{Sample generation} 
Task generation rules are symbolic representations that are handled by KANDY in the form of collections of trees (forests). When rules are parsed and a new tree is generated, it is checked against the set of previously generated symbols (to avoid redundancy) and, optionally, against the ground-truth rule, to guarantee consistency between the generation rules and the ground-truth information. 
Then, the data generator tries to build samples that are inherently different at the {\it symbolic level}, thus strongly emphasizing the symbolic nature of the generation process. Symbol sets are built by sampling trees from either positive or negative generation rules, and grounding their leaves.
Since rules might involve non-deterministic attributes,\footnote{Any value (\texttt{\textasciitilde}), negation (\texttt{not\_*}), and disjunction (\texttt{a|b|\dots|k}).} grounding randomly selects their values.
Finally, perceptual variability is introduced when the image is generated from its symbolic representation, slightly altering size and color. 
%
%
The sampling process is repeated until the target number of examples is produced, unless too many samples are rejected consecutively (controlled via a patience hyper-parameter). In such case, the target size is achieved during rendering by sampling with replacement.
%
%
Symbols are then rendered as images and disjointly split into training, validation and test sets.
Figure~\ref{fig:example-kandy} (bottom) exemplifies generation rules, ground truth, and generated data for two different tasks of a curriculum.


\parafango{Image rendering}
%
%
Images are drawn at a customizable resolution after symbol generation and sample rejection phases. The background color can be customized as well, and shapes can take any color from a discrete set of configurable values.
Similarly, the size of atomic shapes can be controlled by defining a set of sizes of interest, and such a size is not a function of the level of the hierarchy to which atomic shapes belong.\footnote{If the drawing area is too small to draw an atomic shape, overlaps can happen in crowded or highly hierarchical scenes.}
Both size and color can be injected with configurable noise, to implement the already mentioned initial perceptual augmentation.


%

\parafango{Supervisions} KANDY allows to sparsely attach supervisions in the generated curricula of tasks, a process which is modeled by a customizable law $f$. For each task, samples are ordered accordingly to the way the were generated, and the supervision law is function of the sample index $t$, scaled in $[0, 1]$ (i.e, $t=0$ is the first sample of the task and $t=1$ is the last one). 
KANDY implements an exponential decay law which is evaluated for each sample, $f(t) = \gamma \cdot e^{-\sigma \cdot t}$, where $\sigma = \log(\gamma \cdot \beta^{-1})$ and $\gamma, \beta \in [0, 1]$ are hyper-parameters which can be independently defined for each task in the curriculum.
It turns out that function $f(t)$ is monotonically decreasing and has boundary values $f(0) = \gamma$ and $f(1) = \beta$.
Then, a binary random variable is sampled with probability $p_t = f(t)$ and, if the random variable is positive, KANDY marks sample $t$ as supervised, otherwise it is marked as unsupervised.

\parafango{Compositional operators} Compositional primitives are used to model the generation rules (Figure~\ref{fig:example-kandy}, top) that allows KANDY to recursively place and display children objects (either atomic shapes or other compositional primitives) within the image, in function of the bounding boxes of such objects. 
%
%
The simplest operator is called \texttt{in} which draws a child object at the center of the drawing area.
Similarly, operators \texttt{quadrant\_ul, quadrant\_ur, quadrant\_ll, quadrant\_lr}
position the bounding box of an object in one of the four quadrants of the drawing area.
The \texttt{random} operator draws children objects at random positions, using greedy sample rejection to avoid overlapping with previously drawn shapes.
%
Operators \texttt{stack} and \texttt{side\_by\_side} equally split the drawing area available for each child along their primary axis (vertical for \texttt{stack} and horizontal for \texttt{side\_by\_side}).
Objects are drawn left-to-right for \texttt{side\_by\_side} and top-to-bottom for \texttt{stack}.
The variants \texttt{stack\_reduce\_bb} and \texttt{side\_by\_side\_reduce\_bb}, instead, also reduce the secondary axis size. 
The \texttt{diag\_ul\_lr} operator positions children objects from the upper-left corner towards the lower-right, while \texttt{diag\_ll\_ur} arranges childred objects from lower-left to upper-right. 
The \texttt{grid} operator computes the smallest $n$ such that $n^2 \leq \#\mathrm{children}$ 
and children objects are drawn on an $n \times n$ grid from left to right, filling one row at the time, from top to bottom. Grids are incomplete in case the number of children is not a perfect square.

\parafango{Auxiliary operators}
%
%
KANDY also includes additional human-understandable auxiliary operators that simplify the definition of tasks.
These operators are list expansion primitives, which allow for compact definitions, mapping lists into other lists containing only atomic shapes and compositional operators.\footnote{For greater flexibility, list expansions are performed twice, operators whose name includes the \texttt{before} suffix are expanded before grounding the sample tree, while every other operator is expanded after grounding.}
%
In particular, KANDY provides four families of list expansions: ($i.$) deterministic, ($ii.$) random, ($iii.$) memory-based, and ($iv.$) set-based, see also Figure~\ref{fig:example-kandy} (top).
($i.$) Deterministic operators (e.g., sorting or mirroring a list) always produce the same output, for a given input. 
($ii.$) Random operators may produce different results on the same inputs; their behavior can be parameter-free (e.g., random permutation), or controlled by parameters (e.g., random repetition within a range of values).
($iii.$) There are three memory-based operators: \texttt{store} and \texttt{store\_before} return the input list itself, but they have the side effect of memorizing it and associating it to a text alias; \texttt{recall} (which can only be applied after grounding, unless in combination with set operators) takes an alias and returns the list previously associated to it.
($iv.$) Set-based operators have special requirements, as they can only take lists of non-grounded atomic shapes or recalled atomic shapes stored before grounding. They produce a single atomic shape output which is the left-associative application of the set operator to the entire list (e.g., $\texttt{intersect}([\texttt{any triangle}, \texttt{any non red}, \texttt{any large object}]) \mapsto \texttt{any triangle} \cap \texttt{any non red} \cap \texttt{any large} = \texttt{large non red triangle}$). We note that proper care should be taken with intersections, as empty output sets are invalid.



\subsection{Released curricula}\label{sec:curricula}

To show the potential of the KANDY framework and to issue a challenge that could be taken up by several AI communities, 
we release two benchmarks in the form of curricula of tasks, that showcase the features available to the end-user. We name the two curricula {\it easy} and {\it hard}, due to the different complexity of the involved tasks.
In both curricula, each sample is a $224 \times 224$ RGB image annotated with: task ID, binary label, supervision state (whether the label should be used in training or if the sample should be treated as unsupervised), and symbolic representation. 
The symbolic representation is a recursive dictionary of lists representing a tree, i.e., the generation rule for the considered example, as created in the sample generation stage. 
No list expansion operators appear in symbolic annotations.
Atomic objects can be {\it small} or {\it large} ($10 \times 10$ and $25 \times 25$ pixels, respectively), and they can take any of six colors ({\it red}, {\it green}, {\it blue}, {\it cyan}, {\it magenta} and {\it yellow}). Object sizes are corrupted by additive uniform noise in the range $\pm[0, 2]$ pixels, and color is corrupted by zero-mean Gaussian noise in HSV coordinates ($\sigma_H = 0.01, \sigma_S = 0.2, \sigma_V = 0.2$). These values were hand-picked to preserve perceptual boundaries (e.g., humans still perceive the maximally corrupted ``red'' as such). Background was set to gray.

\parafango{Easy} The first curriculum consists of $20$ elementary tasks with limited annotations ($100$ samples for each task, split into $50$ training, $25$ validation, and $25$ test samples).
Tasks $0$ to $8$ are targeted on a single objects and introduce basic (atomic) shapes ($0$-$2$) and color ($3$-$8$). 
Positive instances contain an object possessing some target attributes, while negative ones contain an arbitrary object.
Tasks $9$ to $12$ introduce simple spatial relations between objects. Namely, in such tasks the rightmost object in a positive sample is always a {\it red triangle}, presented in an increasingly complex context: along with an arbitrary object (task $9$), with multiple arbitrary objects ($10$), multiple objects, one of which is a {\it circle} ($11$), and multiple objects, one of which is {\it blue} ($12$). 
Tasks $13$ and $14$ present complex inter-object relations without confounders: positives in task $13$ consist of a {\it triangle} and a {\it square} of the same color, positives in task $14$ are {\it palindromes of three objects} (i.e., A B A displaced horizontally).
Task $15$ to $19$ introduce ``higher order'' objects, which will be reused for the Hard curriculum. These objects are (extremely abstract) representations of: {\it houses} (a square below a triangle), {\it cars} (two side-by-side circles of the same size), {\it towers} (two or three stacked squares of the same size), {\it wagons} (two or three side-by-side squares of the same size), and {\it traffic lights} (red, yellow and green circles of the same size stacked). Negative samples are perceptually similar to positives, but violate the ground truth rules. 

\parafango{Hard} The second curriculum consists of $18$ tasks requiring complex, incremental reasoning capabilities, designed to be challenging for both neural and symbolic methods.
We provide four versions of this curriculum: small fully supervised ($100$ samples per task, $80$ training, $10$ validation, $10$ test samples), large fully supervised ($1000$ total, $800$ train, $100$ validation, $100$ test samples), and two large versions with sparse annotations (both with $1000$ samples, one has fixed $\gamma=\beta=0.5$ probability of supervision, the other has a decay schedule from $\gamma=0.8$ to $\beta=0.2$ probability).
%
Tasks $0$ to $3$ introduce uniform k-mers of two and three objects with a common property. 
Positives in tasks $4$ to $8$ contain the objects introduced at the end of the Easy curriculum, along with confounders (which were not present).
Tasks $9$ to $11$ introduce hierarchical reasoning, with a universally quantified rule applied to each of the groups within a grid. Positives of task $9$ contain a {\it grid} whose elements are displaced in {\it diagonal}, such that it exists a {\it shared shape} in every group (e.g., each contains a square). Likewise, in task $10$ the rule universally quantifies the existence of a color and, in task $11$, the rule involves the complex objects defined in the Easy curriculum (thus introducing a 3-level perceptual hierarchy).
Task $12$ to $14$ extend the {\it palindrome} task of Easy curriculum: in task $12$, positives are a palindrome of arbitrary size between $3$ and $7$ simple objects, displaced along an arbitrary line (horizontal, vertical or diagonal). Task $13$ introduces the concept of ``{\it pseudo-palindrome}'', defined recursively as a sequence $A B A'$ where $B$ is a pseudo-palindrome, and $A$ and $A'$ share either the same shape or the same color. Positives in task $14$ are pseudo-palindromes where couples $A$ and $A'$ can be either simple or complex objects.
%
Positives in task $15$ contain objects of the same color if their number is {\it odd}, or objects of the same shape if their number is {\it even}.
Tasks $16$ and $17$ assess logic implication capabilities on top of complex object recognition. 
Positives in task $16$ contain {\it four objects} satisfying $(\text{traffic light} \Rightarrow \text{car}) \vee (\text{house} \Rightarrow \text{tower})$. Task $17$ universally quantifies task 16 on a grid. 

\section{Experimental evaluation}\label{sec:experiments}
\begin{table}[t]
    \centering
    \begin{minipage}{0.02\textwidth}
    \rotatebox{90}{\hskip -1mm \textbf{Hard-Large} \hskip 1.1cm \textbf{Hard} \hskip 2.7cm \textbf{Easy} \hskip 1.7cm}
    \end{minipage}
    \begin{minipage}{0.7\textwidth}
    \begin{tabular}{l|c|c|c|c}
        \toprule 
        & \footnotesize \textsc{Independent} & \footnotesize \textsc{Joint} & \footnotesize \textsc{Task Incr.} & \footnotesize \textsc{Cont. Online} \\
        \midrule
        \footnotesize \textsc{MLP} & $0.61$ {\tiny ($\pm 0.03$)} & $0.60$ {\tiny ($\pm 0.02$)} & $0.61$ {\tiny ($\pm 0.01$)} & $0.60$ {\tiny ($\pm 0.01$)} \\
        \footnotesize \textsc{CNN} & $\textbf{0.72}$ {\tiny ($\pm 0.00$)} & $\textbf{0.73}$ {\tiny ($\pm 0.01$)} & $\textbf{0.74}$ {\tiny ($\pm 0.02$)} & $0.62$ {\tiny ($\pm 0.01$)} \\
        \footnotesize \textsc{ResNet-50} & $0.66$ {\tiny ($\pm 0.02$)} & $0.67$ {\tiny ($\pm 0.02$)} & $0.61$ {\tiny ($\pm 0.01$)} & $0.53$ {\tiny ($\pm 0.01$)} \\
        \footnotesize \textsc{ResNet-50 (H)} & $\textbf{0.72}$ {\tiny ($\pm 0.02$)} & $0.71$ {\tiny ($\pm 0.02$)} & $0.72$ {\tiny ($\pm 0.02$)} & $\textbf{0.66}$ {\tiny ($\pm 0.01$)} \\
        \footnotesize \textsc{ViT (H)} & $0.69$ {\tiny ($\pm 0.01$)} & $0.70$ {\tiny ($\pm 0.01$)} & $0.70$ {\tiny ($\pm 0.01$)} & $0.60$ {\tiny ($\pm 0.03$)} \\
        [-1mm]\multicolumn{5}{@{}c@{}}{\makebox[1.14\textwidth]{\dashrule}}\\[-1mm]
        \footnotesize \textsc{Aleph (Nat-Min)} \vphantom{$\sqrt{10}^{2^2}$}& $0.91^{\dagger}$ & -- & -- & -- \\
        \footnotesize \textsc{Aleph (Nat-Mid)} & $\textbf{0.96}^{\dagger}$ & -- & -- & -- \\
        \footnotesize \textsc{Aleph (Ptr-Min)} & $0.88^{\dagger}$ & -- & -- & -- \\
        \footnotesize \textsc{Aleph (Ptr-Mid)} & $0.91^{\dagger}$ & -- & -- & -- \\
        \midrule
        \footnotesize \textsc{MLP} & $0.56$ {\tiny ($\pm 0.01$)} & $0.53$ {\tiny ($\pm 0.00$)} & $0.51$ {\tiny ($\pm 0.04$)} & $\textbf{0.55}$ {\tiny ($\pm 0.01$)} \\
        \footnotesize \textsc{CNN} & $0.53$ {\tiny ($\pm 0.01$)} & $\textbf{0.57}$ {\tiny ($\pm 0.02$)} & $0.53$ {\tiny ($\pm 0.02$)} & $0.54$ {\tiny ($\pm 0.02$)} \\
        \footnotesize \textsc{ResNet-50} & $0.52$ {\tiny ($\pm 0.02$)} & $0.51$ {\tiny ($\pm 0.02$)} & $\textbf{0.54}$ {\tiny ($\pm 0.02$)} & $0.52$ {\tiny ($\pm 0.01$)} \\
        \footnotesize \textsc{ResNet-50 (H)} & $\textbf{0.59}$ {\tiny ($\pm 0.03$)} & $\textbf{0.57}$ {\tiny ($\pm 0.02$)} & $0.53$ {\tiny ($\pm 0.02$)} & $0.54$ {\tiny ($\pm 0.01$)} \\
        \footnotesize \textsc{ViT (H)} & $0.52$ {\tiny ($\pm 0.03$)} & $0.51$ {\tiny ($\pm 0.01$)} & $0.51$ {\tiny ($\pm 0.05$)} & $0.54$ {\tiny ($\pm 0.01$)} \\
        [-1mm]\multicolumn{5}{@{}c@{}}{\makebox[1.14\textwidth]{\dashrule}}\\[-1mm]
        \footnotesize \textsc{Aleph (Nat-Min)} \vphantom{$\sqrt{10}^{2^2}$} & $0.65^{\dagger}$ & -- & -- & -- \\
        \footnotesize \textsc{Aleph (Nat-Mid)} & $\textbf{0.66}^{\dagger}$ & -- & -- & -- \\
        \footnotesize \textsc{Aleph (Ptr-Min)} & $0.64^{\dagger}$ & -- & -- & -- \\
        \footnotesize \textsc{Aleph (Ptr-Mid)} & $0.57^{\dagger}$ & -- & -- & -- \\
        \midrule
        \footnotesize \textsc{MLP} & $0.57$ {\tiny ($\pm 0.02$)} & $0.59$ {\tiny ($\pm 0.01$)} & $0.52$ {\tiny ($\pm 0.01$)} & $0.53$ {\tiny ($\pm 0.01$)} \\
        \footnotesize \textsc{CNN} & $0.68$ {\tiny ($\pm 0.00$)} & $\textbf{0.70}$ {\tiny ($\pm 0.01$)} & $0.62$ {\tiny ($\pm 0.01$)} & $0.55$ {\tiny ($\pm 0.01$)} \\
        \footnotesize \textsc{ResNet-50} & $\textbf{0.69}$ {\tiny ($\pm 0.00$)} & $0.61$ {\tiny ($\pm 0.01$)} & $0.54$ {\tiny ($\pm 0.03$)} & $0.55$ {\tiny ($\pm 0.02$)} \\
        \footnotesize \textsc{ResNet-50 (H)} & $0.56$ {\tiny ($\pm 0.01$)} & $0.59$ {\tiny ($\pm 0.00$)} & $0.56$ {\tiny ($\pm 0.01$)} & $0.56$ {\tiny ($\pm 0.01$)} \\
        \footnotesize \textsc{ViT (H)} & $0.63$ {\tiny ($\pm 0.00$)} & $0.65$ {\tiny ($\pm 0.00$)} & $\textbf{0.63}$ {\tiny ($\pm 0.01$)} & $\textbf{0.57}$ {\tiny ($\pm 0.02$)} \\
        \bottomrule
    \end{tabular}
    \end{minipage}
    \begin{minipage}{0.02\textwidth}
    \rotatebox{270}{\hskip 0.7cm \textit{Neural} \hskip 0.5cm \textit{Symbolic} \hskip 0.7cm \textit{Neural} \hskip 0.5cm \textit{Symbolic} \hskip 0.7cm \textit{Neural}}
    \end{minipage}    
    \vskip 3mm
    \caption{\textbf{KANDY-Easy} ({\it top}), \textbf{KANDY-Hard} ({\it middle}) and \textbf{KANDY-Hard-Large} ({\it bottom}), average accuracy on the test data in different learning settings. Results are averaged ($\pm$ standard deviation) over 3 runs (no standard deviation is reported for deterministic models). 
    Results of symbolic methods are marked with $\dagger$ to recall that, for each example, they were given correct perceptual symbolic features.}
    \label{tab:acc_easy_hard}    
\end{table}

\begin{table}[t]
    \centering
    \begin{minipage}{0.02\textwidth}
    \rotatebox{90}{\textbf{Hard} \hskip 0.6cm \textbf{Easy} \hskip 0.4cm}
    \end{minipage}
    \begin{minipage}{0.7\textwidth}
    \begin{tabular}{l|c|c|c|c|c}
        \toprule
        & \footnotesize \sc Correct & \footnotesize \sc Avg Prec & \footnotesize \sc Avg Rec & \footnotesize $C_{gt} / C_{\ell t}$ & \footnotesize $L_{gt} / L_{\ell t}$ \\
        
        \midrule
        \footnotesize \textsc{Aleph (Nat-Min)} & 13/20 & 0.90 & 0.92 & 0.71 & 1.77 \\
        \footnotesize \textsc{Aleph (Nat-Mid)} & 13/20 &  0.94 & 1.0 & 1.0 & 2.39 \\
        \footnotesize \textsc{Aleph (Ptr-Min)} & 13/20 & 0.87 & 0.88 & 0.71 & 0.73 \\
        \footnotesize \textsc{Aleph (Ptr-Mid)} & 13/20 & 0.89 & 0.93 & 0.8 & 0.75 \\
        \midrule
        \footnotesize \textsc{Aleph (Nat-Min)} & 5/18 & 0.66 & 0.47 & 0.19 & 0.49 \\
        \footnotesize \textsc{Aleph (Nat-Mid)} & 6/18 & 0.66 & 0.58 & 0.37 & 0.61 \\
        \footnotesize \textsc{Aleph (Ptr-Min)} & 3/18 & 0.67 & 0.40 & 0.15 & 0.35 \\
        \footnotesize \textsc{Aleph (Ptr-Mid)} & 3/18 & 0.59 & 0.40 & 0.25 & 0.52 \\
        \bottomrule
    \end{tabular}
    \end{minipage}
    \vskip 3mm
    \caption{Metrics describing the theories learned by Aleph. Compression ratios are computed as the number of clauses in the ground truth $C_{gt}$ (literals $L_{gt}$, respectively) divided by those in the learned theory $C_{\ell t}$ (literals $L_{\ell t}$, respectively). A value $< 1.0$ indicates an induced theory larger than the ground truth. $L_{\ell t}$ is computed as the maximum number of literals in any clause in a theory. \vskip -3mm}
    \label{tab:symbolic_results}
\end{table}

The aim of our experiments is to test {\it neural} and {\it symbolic} approaches on the two released curricula of KANDY, described in Section~\ref{sec:curricula} (unless differently specified, we considered the small version of Hard). The goal is to show the limitations of state-of-the-art approaches in both families, thus advocating the need to use NeSy approaches. All the results of this section are about the test splits of the KANDY datasets.


\parafango{Setup} {\it Neural models} are evaluated in solving the task-related binary classification problems. We use the notation $m_j$ to indicate the binary predictor associated to the $j$-th task, which returns outputs in $[0,1]$. The positive class is predicted when $m_j$ exceeds a customizable threshold $\tau_j$.
We evaluated four different learning settings, namely {\small\sc Independent}, {\small\sc Joint}, {\small\sc Task Incremental}, and {\small\sc Continual Online}, the latter two ones are instances of continual learning. 
Models are trained by mini-batch based stochastic optimization (multiple epochs), with the exception of the last learning setting, as detailed in the following:
($1$.) {\small \textsc{Independent}}: each $m_j$ is implemented with its own neural architecture, and it is independently trained using data of the $m$-th task.
($2$.) {\small \textsc{Joint}}: All the $m_j$'s share the same network architecture, except for the output layer, that has an independent head for each $m_j$. This is a classic instance of multi-task learning, where the whole training data (all tasks) is jointly used for learning. 
($3$.) {\small \textsc{Task incremental}}: same network of \textsc{joint}, but tasks are observed incrementally, from simpler to more complex ones, in a continual learning setting. At the $j$-th time step, only the data of the $j$-th task is used for training purposes.
($4$.) {\small \textsc{Continual online}}: same as previous, but exploiting a pure online setting, where the model processes each example only once (single pass, single epoch).
Regarding {\it symbolic models}, we evaluated them in an Inductive Logic Programming (ILP) setting, testing their capability in devising symbolic explanations of the different tasks (in an {\small \textsc{Independent}} manner), comparing with the available ground truth rules.

\parafango{Metrics}
The main metric we consider is {\small\sc Average Accuracy} in $[0,1]$, which is the average of the task-related accuracies $\{\mathrm{acc}_j,\ j=0,\ldots,N-1\}$ ($N$ number of tasks) at the end of learning. In particular, $\mathrm{acc}_{j}$ is the micro-accuracy on the $j$-th  binary classification task.\footnote{A random model yields $\approx 0.50$.} In {\it neural models}, for continual learning settings, where the task data are streamed task-by-task, we overload such definition including a superscript to indicate the ``time'' at which the measurements are made, i.e., $\mathrm{acc}_{j}^{z}$. In this paper, $z$ indicates what is the last task observed by the model. The {\small\sc Average Accuracy} is then evaluated at time $z=N-1$ (i.e., after having processed the last streamed task) when comparing continual settings with non-continual ones. Moreover, we analyze results at different time instants, considering not only accuracies but also other well-established metrics in the continual learning literature~\citep{mai2022online}. The {\small\sc Average Forgetting} at time $z$, for each task $j$, compares $\mathrm{acc}_{j}^{z}$ with the best accuracy obtained in all the previous time instants (i.e., a positive forgetting means the model is performing worse than before); {\small\sc Backward Transfer} measures how what was learned up to time $z$ improved, on average, accuracy in the previous tasks; finally, {\small\sc Forward Transfer} measures how what was learned up to time $t$ improves, on average, accuracy in the not-yet-considered (future) tasks.

\parafango{Neural models} As for our neural baselines, we decided to consider the following models as a representative sample of the most widely employed architectures for image classification:
($i$.) {\small \textsc{MLP}}: multi-layer perceptron with a hidden layer with $100$ neurons (hyperbolic tangent activations), that takes in input a vector of feature where each element is associated to a pixel (spatial information is thus lost).
($ii$.) {\small \textsc{CNN}}: ReLU-based convolutional net, with a stack of $3$ conv. and max pooling layers ($64$ ($5\times5$), $128$ ($3\times3$), $256$ ($3\times3$) features (kernel size), respectively, with stride $2$ in the last two blocks), where the last pooling returns outputs at the resolution of $5\times5$, followed by a dense layer ($4096$ neurons, dropout) and a dense output layer. 
($iii$.) {\small \textsc{ResNet-50}}: ResNet-50 net~\citep{resnet}, trained from scratch.
($iv$.) {\small \textsc{ResNet-50 (H)}}: pre-trained ResNet (ImageNet), re-training the classification head.
($v$.)  {\small \textsc{ViT-16 (H)}}: pre-trained Visual Transformer~\citep{vit}, ViT-Base/16 (ImageNet), where only the classification head is re-trained. 
All models are equipped with sigmoids in the output layer. Training was performed sampling positive/negative training data in a class-balanced manner (for each task). 
Continual learning settings are tackled using experience replay with class-balanced Reservoir Sampling \citep{chrysakis2020online}. The main hyper-parameters where cross-validated by exhaustively searching on fixed grids of values (all the details about the hyper-parameters and configurations are reported in the supplementary material).


\parafango{Symbolic models}
For what concerns symbolic approaches, we rely on the ILP system Aleph, that we train on the Easy curriculum by allowing a runtime of $30$ minutes for each task, and on the small Hard curriculum with an allowed runtime of $1$ hour per task.\footnote{\url{https://www.cs.ox.ac.uk/activities/programinduction/Aleph}}\footnote{We also experimented with Popper (\url{https://github.com/logic-and-learning-lab/Popper}) but the vast majority of our tries timed out for both curricula. The remainder had performance $<$ 0.5 on training data.}
We tested two different encodings, namely natural ({\sc\small Nat}) and pointer (\textsc{\small Ptr}), providing a  
  knowledge base (KB) of two different sizes (\textsc{\small Min}imal and \textsc{\small Mid}-size, respectively).
In both encodings, atomic shapes are constants.
In the \textsc{\small Nat} encoding, compositional operators are unary functors applied to lists. This corresponds to a one-to-one mapping with the symbolic structure used by the KANDY generator, with the advantage of not requiring modifications of the KB.
The \textsc{\small Ptr} encoding, instead, represents compound objects with constants. A hierarchical structure is encoded by a predicate \texttt{defined\_as/3} such as \texttt{defined\_as(obj\_0001, stack, [obj\_0002, triangle\_red\_small, obj\_0003])}, which defines \texttt{obj\_0001} as a pointer to a vertical stacking of three sub-objects (one atomic, and two defined by other \texttt{defined\_as/3} predicates). This encoding has the advantage of not requiring functors, but it splits sample information between KB and example files, requiring an additional \texttt{sample\_is/2} predicate to link them. 
%
As anticipated, we tested two scenarios for what concerns the available knowledge base. In the first one, we only allow a \textsc{\small Min}imal KB consisting of $17$ predicates, which 
is a reasonable starting point for an agnostic learner, aware of the compositional domain, but not of the nature of each task. 
Differently, the \textsc{\small Mid}-size KB contains $39$ predicates. 
This KB is the same used to produce compact ground-truth/rejection rules (maximum $6$ literals for each clause) used at generation time, but it is still insufficient without predicate invention capabilities.\footnote{We also experimented with a \textsc{\small large} KB, that extends \textsc{\small mid-size} with ``cheating'' predicates which encode exactly the ground-truth/rejection rules used at task generation. These additional predicates encapsulate non-Horn metapredicates (such as the swi-prolog \texttt{forall/2} predicate), recursive definitions (such as \texttt{pseudo\_palindrome/1}) and complex objects definitions (such as \texttt{is\_house/2}), in theory allowing for a perfect induction of each task without predicate invention. This last KB is deliberately designed as an upper bound for purely symbolic approaches. However, Aleph fails to produce acceptable solutions for either encoding, as the search space is too large to converge before timeout for most tasks.}
%
Full specifications of the KBs are available in the supplementary material.
All predicates have bounded non-determinacy and can appear in the body of a clause.
%

\subsection{Results on KANDY-Easy}
\begin{figure}
    \centering
    \includegraphics[width=1.0\textwidth]{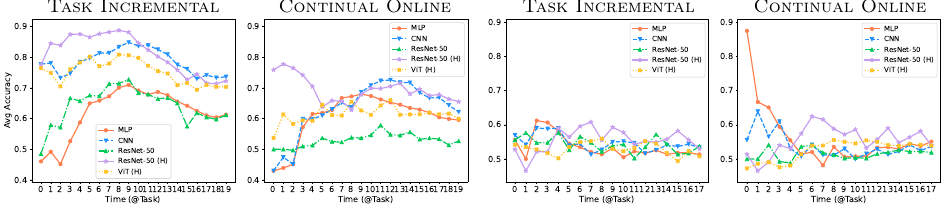}
    \vskip -1mm
    \caption{\textbf{KANDY-Easy} ({\it 1st, 2nd plot}) and \textbf{KANDY-Hard} ({\it 3rd, 4th plot}), average accuracy over time (i.e., after having processed data of each of the sequentially streamed tasks). {\it Left}: task incremental learning. {\it Right}: Continual online learning.}
    \label{fig:time_acc_easy_hard}
\end{figure}

\begin{figure}[!ht]
    \centering
    \includegraphics[width=1.0\textwidth]{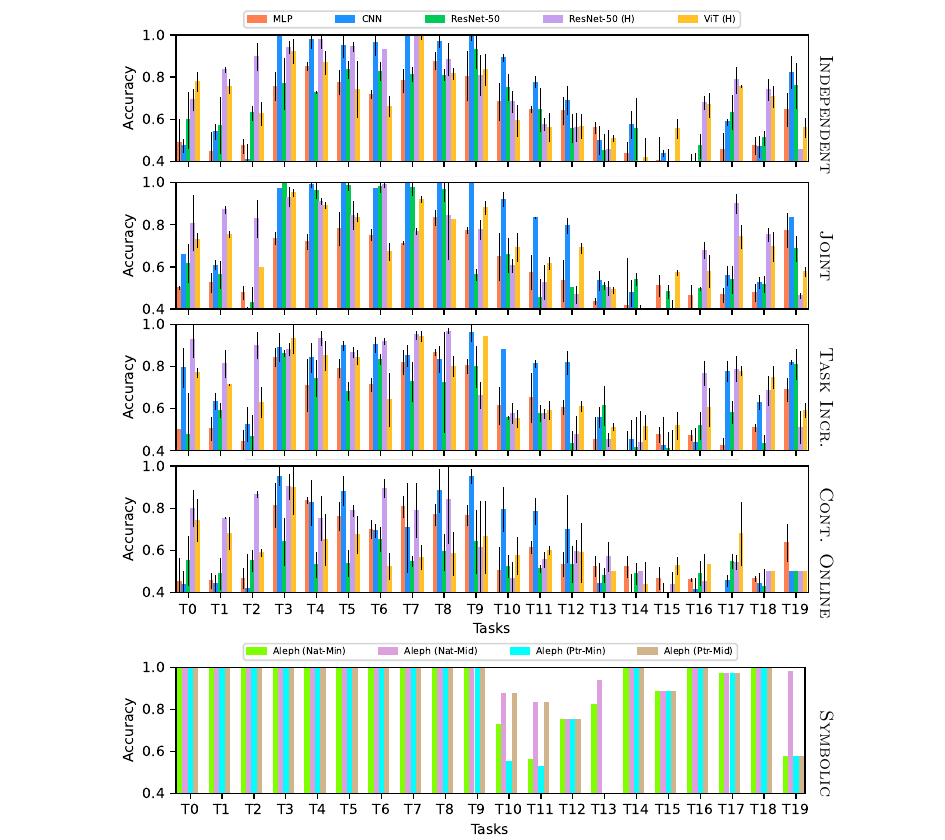}
    \vskip -1mm
    \caption{\textbf{KANDY-Easy}, per-task accuracies of the compared models in different learning settings. Neural (first four rows) and symbolic methods (last row) are represented with different colormaps. Models significantly under-performing (below 0.4) are not shown.}
    \label{fig:acc_per_task_easy}
\end{figure}

\begin{figure}
\centering
    \includegraphics[width=1.0\textwidth]{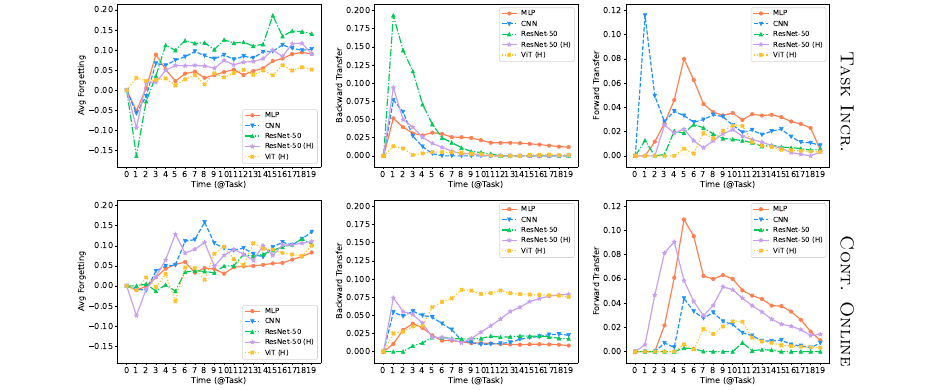}
    \vskip -1mm
    \caption{\textbf{KANDY-Easy}, average forgetting (the lower the better), backward transfer, forward transfer over time (i.e., after having processed data of each of the sequentially streamed tasks). {\it Top}: task incremental learning. {\it Bottom}: Continual online learning.}
    \label{fig:time_easy}
\end{figure}

\parafango{Neural models}
Table~\ref{tab:acc_easy_hard} ({\it top}) reports the {\sc\small Average Accuracy} on test data at the end of learning. Neural nets, overall, reach a max performance of $0.74$, with a preference for the convolutional models, either simple architectures trained from scratch ({\small\sc CNN}) or deeper pre-trained nets ({\small\sc ResNet-50 (H)}). The latter were pre-trained on data that is significantly different from KANDY, but they can still exploit their feature extraction skills to yield improvements with respect to their non-pre-trained counterpart ({\small\sc ResNet-50}). As expected, {\sc\small MLP} more hardly generalizes to the KANDY test data, where shapes can be located in areas that are significantly different from the training ones, and where spatially pooled convolutional models are more appropriate. A too aggressive spatial pooling can also limit the discrimination skills, that is what happens in {\sc\small ResNet-50} (both), where the final pooling yields a $1\times 1$ spatial map, while {\sc\small CNN} pools toward a $5\times 5$ representation which, for example, allows the final fully connected layers to discriminate what is on the left/right/up/bottom side of the image. Overall, {\sc\small ViT (H)} transformers are not far from convolutional models. 
In three out of five cases, training in a {\sc\small Joint} multi-task manner, improves the results, even if with a very small margin. This suggests that the sharing of information among tasks is not strongly promoted, or that the information in each task-data is enough to learn those properties shared among tasks. 
Interestingly, training in a continual manner ({\sc\small Task Incremental}) leads to results that are on par or even better that the other settings, and the temporal dynamics of the accuracy, reported in Figure~\ref{fig:time_acc_easy_hard} ({\it left}), shows somewhat limited drops over time, actually increasing up to half of the time axis (roughly). On one hand, this might either reinforce the hypothesis that there is limited information sharing, and that the network capacity is enough to learn the tasks without much forgetting-prone interference. On the other hand, this might also suggest a different scenario, in which there is indeed a positive sharing of information that does not trigger any forgetting of older, not-shared, information.
To better explore this point, the dynamics of forgetting, forward and backward transfer over time are reported in Figure~\ref{fig:time_easy}, confirming an overall limited, but not negligible, level of forgetting. In particular, during the first time instants we observe positive backward and forward transfers of information, suggesting that the first tasks allow the networks to develop those initial basic skills that are shared with the other tasks, while after a few time instants this does not happen anymore, suggesting a lower sharing of information (in both senses). These considerations open to a nice perspective to study NeSy models trained over time, where the shared knowledge is made explicit. 
The more challenging {\sc\small Continual Online} setting is the one in which there is more room for improvement, since a single-pass on the data yields more difficulties to memorize the basic features of the tasks, as expected (results are $\approx 10\%$ lower than the other setting).
Figure~\ref{fig:acc_per_task_easy} reports per-task accuracies. Overall, the first three tasks are tackled with an accuracy close or above $0.80$ only by a few models, while tasks from $4$ to $10$ are the ones on which many neural models perform well; networks incur in more difficulties in tasks from $11$ to $19$, among which the last three ones look less challenging than the others. Although with different absolute scores, each learning settings shows similar distribution of accuracies. 

\parafango{Symbolic models}
We report in Table~\ref{tab:acc_easy_hard} ({\it top}) the results obtained by Aleph on KANDY-Easy with the different combinations of KBs. We notice how the average test-accuracy is better than those obtained by neural approaches, which is not surprising as Aleph is provided with the correct symbolic representations of each image (as if the perception phase had been perfectly solved). Nevertheless, even in this setting some of the tasks result to be complex, especially with the \textsc{minimal} KB. More details regarding the performance per-task is provided in the supplementary material. We also report in Table~\ref{tab:symbolic_results} additional results describing the learned theories (number of correct clauses, rule precision, recall, compression ratios), to more specifically deepen the quality of such theories. In particular, all the implementations of Aleph are able to learn the correct clause(s) in the ground truth for 13 out of 20 tasks, thus leaving space for improvement. Moreover, we see that the average of per-task recall and per-task precision of the learned rules are both always above 0.85, so that the learned rules (on average) cover almost the total of positive examples, while producing few false positives.
Finally, the compression ratio between the number of clauses in the ground truth $C_{gt}$ (respectively, literals $L_{gt}$) and that in the learned theory $C_{\ell t}$ (respectively, $L_{\ell t}$) is frequently $< 1$, indicating an induced theory that is larger than the ground truth, thus a redundant (suboptimal) solution.

\subsection{Results on KANDY-Hard}
%
%
%

\begin{figure}[!ht]
    \vskip -2mm
    \centering
    \includegraphics[width=1.0\textwidth]{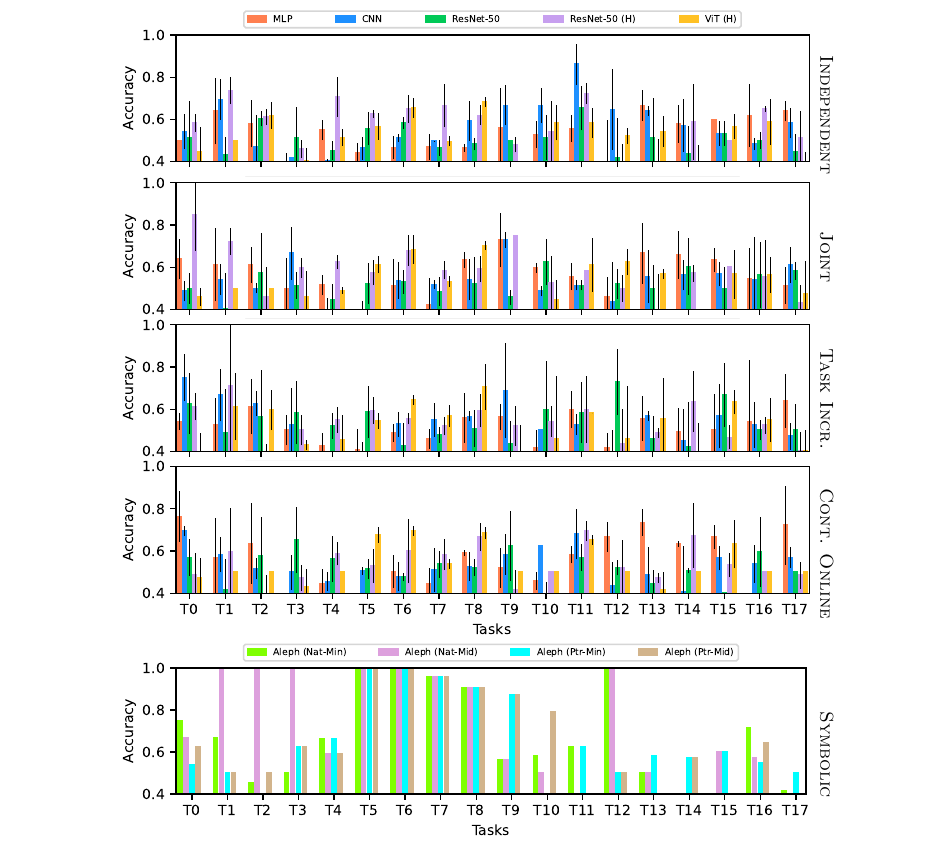}
    \vskip -1mm
    \caption{\textbf{KANDY-Hard}, per-task accuracies of the compared models in different learning settings. Neural (first four rows) and symbolic methods (last row) are represented with different colormaps. Models significantly under-performing (below 0.4) are not shown.\vskip -2mm}
    \label{fig:acc_per_task_hard}
\end{figure}

\begin{figure}
    \centering
    \includegraphics[width=1.0\textwidth]{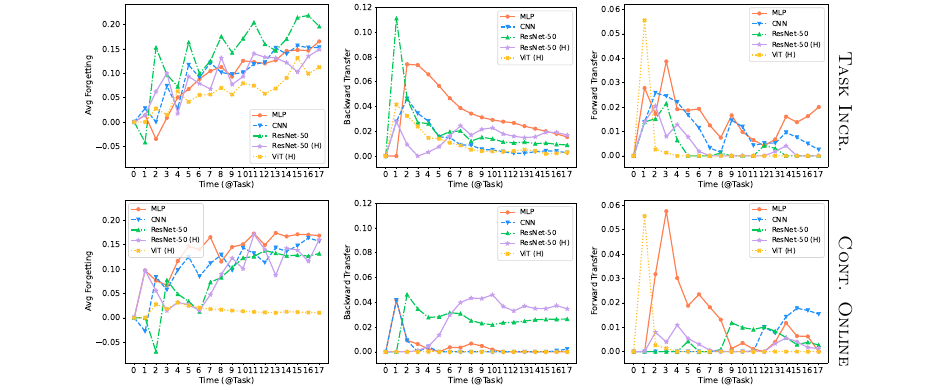}
    \vskip -1mm
    \caption{\textbf{KANDY-Hard}, average forgetting (the lower the better), backward transfer, forward transfer over time (i.e., after having processed data of each of the sequentially streamed tasks). {\it Top}: task incremental learning. {\it Bottom}: Continual online learning.}
    \label{fig:time_hard}
\end{figure}

\parafango{Neural models}
KANDY-Hard is composed of tasks that are significantly harder to solve by neural models, as shown in Table~\ref{tab:acc_easy_hard} ({\it middle}). The top-accuracy is $0.59$, thus not far from a random guess, making it hard to trace strong conclusions on the comparisons among the different models/settings. 
In this case, {\sc\small Continual Online} learning is not different from the multiple-epochs {\sc\small Task Incremental} learning, confirming that multiple passes on the data are not enough to introduce significant improvements. 
Figure~\ref{fig:time_acc_easy_hard} ({\it right}) reports similar temporal dynamics, with the exception of {\sc\small MLP} in {\sc\small Continual Online}, that is able to reach high accuracy on the first task and then incur in evident drops over time. 
The dynamics of forgetting and forward transfer, shown in Figure~\ref{fig:time_hard}, are similar to KANDY-Easy. {\sc\small ViT (H)} seems to not forget, but it is due to accuracy scores too close to the random guess. The backward transfer is more evident on the long run, compared to KANDY-Easy, while it is less effective in the first time instants, showing an intrinsically more complicated capability of sharing information, due to the difficulty of the tasks, leaving room to novel research.
Per-task accuracies are reported in Figure~\ref{fig:acc_per_task_hard}. Tasks $1$, $2$, $9$, and $11$ are the ones on which neural models perform slightly better. 
We performed additional experiments with a larger version of the Hard dataset ($1000$ samples per task, fully supervised), reporting results in Table~\ref{tab:acc_easy_hard} ({\it bottom}), to assess whether there is room for improvement with additional data. We observe improvements in terms of accuracy, that are coherent with the overall picture discussed on KANDY-Easy. This confirms that the benchmark is very challenging and neural models in continual learning settings struggle more than in the Easy case, opening to several possible investigations with (continual) NeSy methods. 

\parafango{Symbolic models}
Table~\ref{tab:acc_easy_hard} ({\it middle}) reports the results obtained by Aleph on KANDY-Hard. Even though Aleph is given the correct symbolic representations of each image, most of the tasks still remain extremely challenging. The statistics reported in Table~\ref{tab:symbolic_results} ({\it bottom}) highlight that only for a few (3 to 6) of the 18 tasks, the correct solution is induced. The learned theories typically contain many clauses (low ratio $C_{gt} / C_{\ell t}$) specialized and covering few cases (higher precision than recall), and manual inspection (see supplementary material) revealed the presence of ``overfitting'' facts in the form \texttt{valid(training\_sample)}, which inflate the $C_{gt} / C_{\ell t}$ ratio without improving performance. Slightly better performance can be observed with \textsc{Nat} encoding rather than \textsc{Ptr}, and with \textsc{Mid} KB with respect to \textsc{Min}: the number of correct tasks is higher and both compression ratios are closer to 1. 
While this can be in part attributed to encodings other than \textsc{Nat-Mid} requiring longer clauses for the same concept, low compression ratios on \textsc{Nat-Mid} hint at a generalized difficulty in inducing effective clauses. Overall, these results confirm that KANDY-Hard is extremely challenging even for symbolic methods with a non-noisy KB: this calls for the application of methods 
combining complex hierarchical feature extraction with strong reasoning.

%

\section{Conclusions}\label{sec:conclusions}
We presented KANDY, a Kandinsky pattern inspired benchmark and dataset-generation tool for the novel setting of Incremental Neuro-Symbolic Learning and Reasoning.
KANDY can be used to generate fully customizable sequences of binary classification tasks organized into a curriculum-like structure (simpler concepts are presented before harder ones), and requiring first-order logic reasoning capabilities.
Samples in KANDY are simple geometric images possessing highly-hierarchical structure, which can be mapped to a symbolic representation (a tree of compositions and shapes). 
A supervision decay schedule can be applied to labels for the investigation of semi-supervised settings, and the presence of symbolic annotations allows to diagnose separately perceptual (from image to symbol) and reasoning (from symbol to decision) components.
%
Results on baseline neural and purely symbolic baselines highlight margin for improvement by means of integrated learning and reasoning methods. We hope that the curricular nature of KANDY will spark cooperation between the Neuro-Symbolic, Continual Learning and Semi-Supervised Learning communities.

\backmatter

\bmhead{Supplementary Material}

We provide a supplementary document with several details on KANDY and on the experimental setup.

%

\section*{Declarations}
\vskip -1mm
\begin{itemize}
\item {\it Funding:} This work was partially supported by the TAILOR project, funded by EU Horizon 2020 research and innovation programme under GA No 952215.
\item {\it Conflict of interest/Competing interests:} None.
\item {\it Ethics approval:} Not applicable.
\item {\it Consent to participate:} Not applicable.
\item {\it Consent for publication:} Not applicable.
\item {\it Availability of data and materials:} Data can be freely downloaded at \url{https://github.com/continual-nesy/KANDYBenchmark/releases/latest}
\item {\it Code availability:} Code can be freely downloaded at \url{https://github.com/continual-nesy/KANDYBenchmark}
\item {\it Authors' contributions:} L.S.L. performed symbolic experiments.
M.L. contributed with literature review and metrics evaluation.
S.M. ran neural experiments and prepared figures and tables. All authors contributed equally to the ideas and writing.
\end{itemize}

%

\setlength{\bibsep}{1.0ex} 
\bibliography{biblio}

\begin{thebibliography}{36}
\providecommand{\natexlab}[1]{#1}
\providecommand{\url}[1]{{#1}}
\providecommand{\urlprefix}{URL }
\providecommand{\doi}[1]{\url{https://doi.org/#1}}
\providecommand{\eprint}[2][]{\url{#2}}
 \bibcommenthead

\bibitem[{Ai et~al(2023)Ai, Langer, Muggleton, and Schmid}]{ai2023explanatory}
Ai L, Langer J, Muggleton SH, et~al (2023) Explanatory machine learning for sequential human teaching. Machine Learning 112(10):3591--3632

\bibitem[{Barrett et~al(2018)Barrett, Hill, Santoro, Morcos, and Lillicrap}]{barrett2018measuring}
Barrett D, Hill F, Santoro A, et~al (2018) Measuring abstract reasoning in neural networks. In: International conference on machine learning, PMLR, pp 511--520

\bibitem[{Benny et~al(2021)Benny, Pekar, and Wolf}]{benny2021scale}
Benny Y, Pekar N, Wolf L (2021) Scale-localized abstract reasoning. In: Proc. of the IEEE/CVF Conf. on Comp. Vis. and Patt. Rec., pp 12557--12565

\bibitem[{Bongard(1970)}]{bongard1970pattern}
Bongard N (1970) Pattern recognition. Spartan Books, New York

\bibitem[{Chollet(2019)}]{chollet2019measure}
Chollet F (2019) On the measure of intelligence. arXiv:191101547

\bibitem[{Chrysakis and Moens(2020)}]{chrysakis2020online}
Chrysakis A, Moens MF (2020) Online continual learning from imbalanced data. In: International Conference on Machine Learning, PMLR, pp 1952--1961

\bibitem[{Dosovitskiy et~al(2021)Dosovitskiy, Beyer, Kolesnikov, Weissenborn, Zhai, Unterthiner, Dehghani, Minderer, Heigold, Gelly, Uszkoreit, and Houlsby}]{vit}
Dosovitskiy A, Beyer L, Kolesnikov A, et~al (2021) An image is worth 16x16 words: Transformers for image recognition at scale. In: Int. Conf. on Learn. Represent.

\bibitem[{He et~al(2016)He, Zhang, Ren, and Sun}]{resnet}
He K, Zhang X, Ren S, et~al (2016) Deep residual learning for image recognition. In: 2016 IEEE Conf. on Computer Vision and Pattern Rec., pp 770--778

\bibitem[{Helff et~al(2023)Helff, Stammer, Shindo, Dhami, and Kersting}]{helff2023v}
Helff L, Stammer W, Shindo H, et~al (2023) V-lol: A diagnostic dataset for visual logical learning. arXiv preprint arXiv:230607743

\bibitem[{Hersche et~al(2023)Hersche, Zeqiri, Benini, Sebastian, and Rahimi}]{hersche2023neuro}
Hersche M, Zeqiri M, Benini L, et~al (2023) A neuro-vector-symbolic architecture for solving raven’s progressive matrices. Nature Machine Intelligence 5(4):363--375

\bibitem[{Holzinger et~al(2019)Holzinger, Kickmeier-Rust, and M{\"u}ller}]{holzinger2019kandinsky}
Holzinger A, Kickmeier-Rust M, M{\"u}ller H (2019) Kandinsky patterns as iq-test for machine learning. In: Machine Learning and Knowledge Extraction: 3rd International Cross-Domain Conference, CD-MAKE, Springer, pp 1--14

\bibitem[{Holzinger et~al(2021)Holzinger, Saranti, and Mueller}]{holzinger2021kandinskypatterns}
Holzinger A, Saranti A, Mueller H (2021) {KANDINSKYPatterns}--{An experimental exploration environment for Pattern Analysis and Machine Intelligence}. arXiv preprint arXiv:210300519

\bibitem[{Hu et~al(2021)Hu, Ma, Liu, Wei, and Bai}]{hu2021stratified}
Hu S, Ma Y, Liu X, et~al (2021) Stratified rule-aware network for abstract visual reasoning. In: Proc. of the AAAI Conference on Artificial Intelligence, pp 1567--1574

\bibitem[{Jiang et~al(2023)Jiang, Tang, Li, and Liu}]{jiang2023bongard}
Jiang G, Tang C, Li Y, et~al (2023) Bongard-tool: Tool concept induction from few-shot visual exemplars. In: PKU 22Fall Course: Cognitive Reasoning

\bibitem[{Jiang et~al(2022)Jiang, Ma, Nie, Yu, Zhu, and Anandkumar}]{jiang2022bongard}
Jiang H, Ma X, Nie W, et~al (2022) Bongard-hoi: Benchmarking few-shot visual reasoning for human-object interactions. In: IEEE/CVF Conf. on Computer Vision and Pattern Rec., pp 19056--19065

\bibitem[{Johnson et~al(2017)Johnson, Hariharan, Van Der~Maaten, Fei-Fei, Lawrence~Zitnick, and Girshick}]{johnson2017clevr}
Johnson J, Hariharan B, Van Der~Maaten L, et~al (2017) Clevr: A diagnostic dataset for compositional language and elementary visual reasoning. In: Proc. of the IEEE conference on computer vision and pattern recognition, pp 2901--2910

\bibitem[{Kaur et~al(2022)Kaur, Uslu, Rittichier, and Durresi}]{kaur2022trustworthy}
Kaur D, Uslu S, Rittichier KJ, et~al (2022) Trustworthy artificial intelligence: a review. ACM Computing Surveys (CSUR) 55(2):1--38

\bibitem[{Larson and Michalski(1977)}]{larson1977inductive}
Larson J, Michalski RS (1977) Inductive inference of vl decision rules. ACM SIGART Bulletin (63):38--44

\bibitem[{Liu et~al(2023)Liu, Lu, and Mou}]{liu2023weakly}
Liu X, Lu Z, Mou L (2023) Weakly supervised reasoning by neuro-symbolic approaches. arXiv preprint arXiv:230913072

\bibitem[{Mai et~al(2022)Mai, Li, Jeong, Quispe, Kim, and Sanner}]{mai2022online}
Mai Z, Li R, Jeong J, et~al (2022) Online continual learning in image classification: An empirical survey. Neurocomputing 469:28--51

\bibitem[{Marconato et~al(2023)Marconato, Bontempo, Ficarra, Calderara, Passerini, and Teso}]{DBLP:conf/icml/MarconatoBFCPT23}
Marconato E, Bontempo G, Ficarra E, et~al (2023) Neuro-symbolic continual learning: Knowledge, reasoning shortcuts and concept rehearsal. In: Proc. of the International Conference on Machine Learning, {ICML}, pp 23915--23936

\bibitem[{Michalski(1980)}]{michalski1980pattern}
Michalski RS (1980) Pattern recognition as rule-guided inductive inference. IEEE Transactions on Pattern Analysis and Machine Intelligence (4):349--361

\bibitem[{M{\"u}ller and Holzinger(2021)}]{muller2021kandinsky}
M{\"u}ller H, Holzinger A (2021) Kandinsky patterns. Artificial intelligence 300:103546

\bibitem[{Nie et~al(2020)Nie, Yu, Mao, Patel, Zhu, and Anandkumar}]{nie2020bongard}
Nie W, Yu Z, Mao L, et~al (2020) Bongard-logo: A new benchmark for human-level concept learning and reasoning. Advances in Neur Inf Proc Sys 33:16468--16480

\bibitem[{Ott et~al(2023)Ott, Ledaguenel, Hudelot, and Hartwig}]{ott2023think}
Ott J, Ledaguenel A, Hudelot C, et~al (2023) How to think about benchmarking neurosymbolic ai? In: 17th Int. Work. on Neural-Symbolic Learning and Reasoning

\bibitem[{Raven(1938)}]{raven1938raven}
Raven JC (1938) Raven standard progressive matrices. Journ of Cognit and Devel

\bibitem[{Shindo et~al(2023)Shindo, Pfanschilling, Dhami, and Kersting}]{shindo2023alpha}
Shindo H, Pfanschilling V, Dhami DS, et~al (2023) $\alpha$ {ILP}: thinking visual scenes as differentiable logic programs. Machine Learning 112(5):1465--1497

\bibitem[{Spratley et~al(2023)Spratley, Ehinger, and Miller}]{spratley2023unicode}
Spratley S, Ehinger KA, Miller T (2023) Unicode analogies: An anti-objectivist visual reasoning challenge. In: Proc. of the IEEE/CVF Conf. on Comp. Vis. and Patt. Rec., pp 19082--19091

\bibitem[{Stammer et~al(2022)Stammer, Memmel, Schramowski, and Kersting}]{stammer2022interactive}
Stammer W, Memmel M, Schramowski P, et~al (2022) Interactive disentanglement: Learning concepts by interacting with their prototype representations. In: IEEE Conf. on Computer Vision and Pattern Rec., pp 10317--10328

\bibitem[{Teney et~al(2020)Teney, Wang, Cao, Liu, Shen, and van~den Hengel}]{teney2020v}
Teney D, Wang P, Cao J, et~al (2020) V-prom: A benchmark for visual reasoning using visual progressive matrices. In: Proc. of the AAAI Conference on Artificial Intelligence, pp 12071--12078

\bibitem[{Treisman(1998)}]{treisman1998feature}
Treisman A (1998) Feature binding, attention and object perception. Philos Trans of the Royal Society of London Series B: Biological Sciences 353(1373):1295--1306

\bibitem[{Wang et~al(2021)Wang, Chen, and Zhu}]{wang2021survey}
Wang X, Chen Y, Zhu W (2021) A survey on curriculum learning. IEEE Transactions on Pattern Analysis and Machine Intelligence 44(9):4555--4576

\bibitem[{Yin et~al(2022)Yin, Lu, Wang, You, Zhang, Wang, Zhen, and Wan}]{yin2022visual}
Yin A, Lu W, Wang S, et~al (2022) Visual perception inference on raven’s progressive matrices by semi-supervised contrastive learning. In: CAAI International Conference on Artificial Intelligence, Springer, pp 399--412

\bibitem[{Youssef et~al(2022)Youssef, Ze{\v{c}}evi{\'c}, Dhami, and Kersting}]{youssef2022towards}
Youssef S, Ze{\v{c}}evi{\'c} M, Dhami DS, et~al (2022) Towards a solution to bongard problems: A causal approach. arXiv preprint arXiv:220607196

\bibitem[{Yun et~al(2020)Yun, Bohn, and Ling}]{yun2020deeper}
Yun X, Bohn T, Ling C (2020) A deeper look at bongard problems. In: Advances in Artificial Intelligence: 33rd Canadian AI 2020, Springer, pp 528--539

\bibitem[{Zhang et~al(2019)Zhang, Gao, Jia, Zhu, and Zhu}]{zhang2019raven}
Zhang C, Gao F, Jia B, et~al (2019) Raven: A dataset for relational and analogical visual reasoning. In: Proc. of the IEEE/CVF Conf. on Comp. Vis. and Patt. Rec.

\end{thebibliography}



\begin{thebibliography}{1}
\ifx \bisbn   \undefined \def \bisbn  #1{ISBN #1}\fi
\ifx \binits  \undefined \def \binits#1{#1}\fi
\ifx \bauthor  \undefined \def \bauthor#1{#1}\fi
\ifx \batitle  \undefined \def \batitle#1{#1}\fi
\ifx \bjtitle  \undefined \def \bjtitle#1{#1}\fi
\ifx \bvolume  \undefined \def \bvolume#1{\textbf{#1}}\fi
\ifx \byear  \undefined \def \byear#1{#1}\fi
\ifx \bissue  \undefined \def \bissue#1{#1}\fi
\ifx \bfpage  \undefined \def \bfpage#1{#1}\fi
\ifx \blpage  \undefined \def \blpage #1{#1}\fi
\ifx \burl  \undefined \def \burl#1{\textsf{#1}}\fi
\ifx \doiurl  \undefined \def \doiurl#1{\url{https://doi.org/#1}}\fi
\ifx \betal  \undefined \def \betal{\textit{et al.}}\fi
\ifx \binstitute  \undefined \def \binstitute#1{#1}\fi
\ifx \binstitutionaled  \undefined \def \binstitutionaled#1{#1}\fi
\ifx \bctitle  \undefined \def \bctitle#1{#1}\fi
\ifx \beditor  \undefined \def \beditor#1{#1}\fi
\ifx \bpublisher  \undefined \def \bpublisher#1{#1}\fi
\ifx \bbtitle  \undefined \def \bbtitle#1{#1}\fi
\ifx \bedition  \undefined \def \bedition#1{#1}\fi
\ifx \bseriesno  \undefined \def \bseriesno#1{#1}\fi
\ifx \blocation  \undefined \def \blocation#1{#1}\fi
\ifx \bsertitle  \undefined \def \bsertitle#1{#1}\fi
\ifx \bsnm \undefined \def \bsnm#1{#1}\fi
\ifx \bsuffix \undefined \def \bsuffix#1{#1}\fi
\ifx \bparticle \undefined \def \bparticle#1{#1}\fi
\ifx \barticle \undefined \def \barticle#1{#1}\fi
\bibcommenthead
\ifx \bconfdate \undefined \def \bconfdate #1{#1}\fi
\ifx \botherref \undefined \def \botherref #1{#1}\fi
\ifx \url \undefined \def \url#1{\textsf{#1}}\fi
\ifx \bchapter \undefined \def \bchapter#1{#1}\fi
\ifx \bbook \undefined \def \bbook#1{#1}\fi
\ifx \bcomment \undefined \def \bcomment#1{#1}\fi
\ifx \oauthor \undefined \def \oauthor#1{#1}\fi
\ifx \citeauthoryear \undefined \def \citeauthoryear#1{#1}\fi
\ifx \endbibitem  \undefined \def \endbibitem {}\fi
\ifx \bconflocation  \undefined \def \bconflocation#1{#1}\fi
\ifx \arxivurl  \undefined \def \arxivurl#1{\textsf{#1}}\fi
\csname PreBibitemsHook\endcsname

\bibitem[\protect\citeauthoryear{Mai et~al.}{2022}]{mai2022online}
\begin{barticle}
\bauthor{\bsnm{Mai}, \binits{Z.}},
\bauthor{\bsnm{Li}, \binits{R.}},
\bauthor{\bsnm{Jeong}, \binits{J.}},
\bauthor{\bsnm{Quispe}, \binits{D.}},
\bauthor{\bsnm{Kim}, \binits{H.}},
\bauthor{\bsnm{Sanner}, \binits{S.}}:
\batitle{Online continual learning in image classification: An empirical survey}.
\bjtitle{Neurocomputing}
\bvolume{469},
\bfpage{28}--\blpage{51}
(\byear{2022})
\end{barticle}
\endbibitem

\end{thebibliography}

\end{document}


\title[The KANDY Benchmark - Supplementary]{The KANDY Benchmark: Incremental Neuro-Symbolic Learning and Reasoning with Kandinsky Patterns -- {\bf Supplementary Material}}


\author*[1,2]{\fnm{Luca Salvatore} \sur{Lorello}}\email{luca.lorello@phd.unipi.it}

\author[2]{\fnm{Marco} \sur{Lippi}}\email{marco.lippi@unimore.it}

\author[3]{\fnm{Stefano} \sur{Melacci}}\email{mela@diism.unisi.it}

\affil[1]{\orgdiv{Department of Computer Science}, \orgname{University of Pisa}, \orgaddress{\street{Largo B.~Pontecorvo 3}, \city{Pisa}, \postcode{56127}, \country{Italy}}}

\affil[2]{\orgdiv{Department of Sciences and Methods for Engineering}, \orgname{University of Modena and Reggio Emilia}, \orgaddress{\street{via Amendola 2}, \city{Reggio Emilia}, \postcode{42122}, \country{Italy}}}

\affil[3]{\orgdiv{Department of Information Engineering and Mathematics}, \orgname{University of Siena}, \orgaddress{\street{via Roma 56}, \city{Siena}, \postcode{53100}, \country{Italy}}}

\maketitle

\section{Overview}
This supplementary document includes several further details on different aspects of our experiments (Section~\ref{sec:appendix_hyper}, Section~\ref{sec:appendix_metrics}) and of the KANDY generation pipeline, including examples of full task descriptions, generated images, background knowledge (Section~\ref{uff}). Moreover, the details on knowledge bases used in experiments with symbolic models are included as well.

\section{Cross-validation and optimal values of the hyper-parameters}
\label{sec:appendix_hyper}
Each task in the KANDY datasets comes with a training, validation and test split. The validation splits were used to measure the average accuracy (further averaged over $3$ runs with different initialization, in the case of neural models) of each target model, varying the values of a number of key hyper-parameters over fixed grids.

\parafango{Neural models} In the case of neural models, we provided data in mini-batches of size $b \in \{1, 16\}$, being $\mathcal{B}$ a randomly sampled mini-batch with $|\mathcal{B}|=b$, exploiting the Adam optimizer or plain Stochastic Gradient Descent (SGD). The initial learning rate ($\mathtt{lr}$) of Adam and the learning rate of SGD were selected from $\{10^{-4}, 10^{-2}\}$, and the number of epochs ($\mathtt{ep}$) was set to a value in $\{1, 10\}$. Notice that in {\small\sc Continual Online} learning, the number of epochs is always $1$ (since it is a single-pass setting), while in the other settings it represents the overall number of training epochs ({\small\sc Joint}) or the number of training epochs per task ({\small\sc Independent}, {\small\sc Task Incremental}). In {\small\sc Task Incremental} and {\small\sc Continual Online}, we tested both the case in which no-experiences are replayed (replay buffer $\mathcal{R}$ size is $0$) and the case in which a replay buffer of size $200$ is used ($|\mathcal{R}|=200$). In the latter case, we set $\lambda_r$ to  $1.0$, being it the coefficient that weigh the contribution of the loss that is about the replayed experiences. In detail, if $\mathcal{L}$ is the binary cross entropy loss (the one we used) and a $(x,y)$ is a pair composed of an input sample and its positive/negative label, respectively, then the overall loss we considered in our stochastic-optimization-based experience is:
$$
\frac{1}{b} \sum_{\substack{x_i \in \mathcal{B}\\ i=1}}^{b} \mathcal{L}(f_{\texttt{task}(x_i)}(x_i), y_i) + \lambda_r \frac{1}{b} \sum_{\substack{x_j \in \tilde{\mathcal{R}}\\ j=1}}^{b} \mathcal{L}(f_{\texttt{task}(x_j)}(x_j), y_j),
$$
where $\tilde{\mathcal{R}}\subset \mathcal{R}$ is a mini-batch which is randomly sampled from the replay buffer $\mathcal{R}$ and $\texttt{task}(x)$ is the function that returns the task index of example $x$. Of course, in non-continual settings, $\lambda_r$ is forced to be $0$.
No data augmentation was used, since we found it to not provide significant help after having run an initial set of experiments.
In Table~\ref{tab:hyper} (first and second column) we report the optimal values of the hyper-parameters that were cross-validated during our experimental experience, considering the KANDY-Easy and KANDY-Hard datasets, respectively. Table~\ref{tab:hyper} (third column) also reports the optimal hyper-parameter values in the case of KANDY-Hard-Large (i.e., the one with increased data/supervisions).
In KANDY-Easy (Table~\ref{tab:hyper}, first column) we can easily see a strong tendency in preferring the Adam optimizer, multiple epochs, and larger batch sizes, with the exception of {\small\sc MLP}, where SGD, one epoch, and a small batch size were found to perform better. This is due to the intrinsic difficulty that the model has in finding a good trade-off between fitting the training data and generalizing to never-seen-before examples, since test data might contain shapes that are located in areas of the input image that are significantly different from the training ones, a condition that can hardly be handled by a vanilla {\small\sc MLP}. As a consequence, multiple epochs (and more informed/specifically-tuned updates due to larger batches and Adam) turned out to reduce the out-of-sample generalization skills, specializing more on the training instances. Not surprisingly, all the models exploit a non-null buffer size in the continual learning settings, with the exception of the already mentioned {\small\sc MLP} case (for the same reasons). In KANDY-Hard (Table~\ref{tab:hyper}, second column), the just-described scenario propagates to more models, due to the intrinsic larger difficulty of the tasks. Interestingly, increasing the amount of supervised examples (Table~\ref{tab:hyper}, third column) reduces such a tendency, since the more informative training data allows the model to have better chances to find test examples that are more closely related to training ones. In this case, multiple epochs are always preferred, and larger batch sizes are selected most of the times.

\begin{table}[t]
    \centering
    \begin{tabular}{l@{\hspace{1mm}}|p{0.5cm}p{0.5cm}p{0.05cm}p{0.05cm}p{0.3cm}|p{0.5cm}p{0.5cm}p{0.05cm}p{0.05cm}p{0.3cm}|p{0.5cm}p{0.5cm}p{0.05cm}p{0.05cm}p{0.3cm}}
        \multicolumn{1}{c}{$\ $} & \multicolumn{5}{c}{\textbf{Easy}} & \multicolumn{5}{c}{\textbf{Hard}} & \multicolumn{5}{c}{\textbf{Hard-Large}} \\
        \toprule
        & \multicolumn{5}{|c|}{\footnotesize \textsc{Independent}} & \multicolumn{5}{|c|}{\footnotesize \textsc{Independent}} & \multicolumn{5}{|c}{\footnotesize \textsc{Independent}} \\
        & \footnotesize $\mathtt{lr}$ & \footnotesize $\mathtt{optim}$ & \footnotesize $b$ & \footnotesize $\mathtt{ep}$ & \footnotesize $|\mathcal{R}|$ & \footnotesize $\mathtt{lr}$ & \footnotesize $\mathtt{optim}$ & \footnotesize $b$ & \footnotesize $\mathtt{ep}$ & \footnotesize $|\mathcal{R}|$ & \footnotesize $\mathtt{lr}$ & \footnotesize $\mathtt{optim}$ & \footnotesize $b$ & \footnotesize $\mathtt{ep}$ & \footnotesize $|\mathcal{R}|$ \\
        \midrule
        \footnotesize \textsc{MLP} & 0.01 & SGD & 1 & 1 & 0 & 0.0001 & Adam & 16 & 1 & 0 & 0.0001 & Adam & 16 & 10 & 0 \\
        \footnotesize \textsc{CNN} & 0.01 & SGD & 1 & 10 & 0 & 0.01 & SGD & 16 & 10 & 0 & 0.0001 & Adam & 16 & 10 & 0 \\
        \footnotesize \textsc{ResNet-50} & 0.0001 & Adam & 16 & 10 & 0 & 0.0001 & Adam & 16 & 10 & 0 & 0.0001 & Adam & 16 & 10 & 0 \\
        \footnotesize \textsc{ResNet-50 (H)} & 0.01 & Adam & 16 & 10 & 0 & 0.01 & Adam & 16 & 10 & 0 & 0.01 & Adam & 16 & 10 & 0 \\
        \footnotesize \textsc{ViT (H)} & 0.01 & Adam & 16 & 10 & 0 & 0.0001 & Adam & 1 & 10 & 0 & 0.01 & SGD & 1 & 10 & 0 \\
        \midrule
        & \multicolumn{5}{|c|}{\footnotesize \textsc{Joint}} & \multicolumn{5}{|c|}{\footnotesize \textsc{Joint}} & \multicolumn{5}{|c}{\footnotesize \textsc{Joint}} \\
        & \footnotesize $\mathtt{lr}$ & \footnotesize $\mathtt{optim}$ & \footnotesize $b$ & \footnotesize $\mathtt{ep}$ & \footnotesize $|\mathcal{R}|$ & \footnotesize $\mathtt{lr}$ & \footnotesize $\mathtt{optim}$ & \footnotesize $b$ & \footnotesize $\mathtt{ep}$ & \footnotesize $|\mathcal{R}|$ & \footnotesize $\mathtt{lr}$ & \footnotesize $\mathtt{optim}$ & \footnotesize $b$ & \footnotesize $\mathtt{ep}$ & \footnotesize $|\mathcal{R}|$ \\
        \midrule
        \footnotesize \textsc{MLP} & 0.01 & SGD & 1 & 1 & 0 & 0.0001 & SGD & 1 & 1 & 0 & 0.0001 & Adam & 1 & 10 & 0 \\
        \footnotesize \textsc{CNN} & 0.0001 & Adam & 1 & 10 & 0 & 0.01 & SGD & 16 & 10 & 0 & 0.0001 & Adam & 1 & 10 & 0 \\
        \footnotesize \textsc{ResNet-50} & 0.01 & Adam & 16 & 10 & 0 & 0.0001 & Adam & 1 & 10 & 0 & 0.0001 & Adam & 16 & 10 & 0 \\
        \footnotesize \textsc{ResNet-50 (H)} & 0.01 & Adam & 16 & 10 & 0 & 0.01 & Adam & 16 & 10 & 0 & 0.01 & Adam & 16 & 10 & 0 \\
        \footnotesize \textsc{ViT (H)} & 0.01 & Adam & 16 & 10 & 0 & 0.01 & SGD & 1 & 10 & 0 & 0.01 & Adam & 16 & 10 & 0 \\
        \midrule
        & \multicolumn{5}{|c|}{\footnotesize \textsc{Task Incremental}} & \multicolumn{5}{|c|}{\footnotesize \textsc{Task Incremental}} & \multicolumn{5}{|c}{\footnotesize \textsc{Task Incremental}} \\
        & \footnotesize $\mathtt{lr}$ & \footnotesize $\mathtt{optim}$ & \footnotesize $b$ & \footnotesize $\mathtt{ep}$ & \footnotesize $|\mathcal{R}|$ & \footnotesize $\mathtt{lr}$ & \footnotesize $\mathtt{optim}$ & \footnotesize $b$ & \footnotesize $\mathtt{ep}$ & \footnotesize $|\mathcal{R}|$ & \footnotesize $\mathtt{lr}$ & \footnotesize $\mathtt{optim}$ & \footnotesize $b$ & \footnotesize $\mathtt{ep}$ & \footnotesize $|\mathcal{R}|$ \\
        \midrule
        \footnotesize \textsc{MLP} & 0.01 & SGD & 1 & 1 & 0 & 0.01 & SGD & 1 & 1 & 0 & 0.0001 & Adam & 1 & 10 & 0 \\
        \footnotesize \textsc{CNN} & 0.0001 & Adam & 1 & 10 & 200 & 0.0001 & Adam & 1 & 1 & 200 & 0.01 & SGD & 16 & 10 & 200 \\
        \footnotesize \textsc{ResNet-50} & 0.01 & Adam & 16 & 10 & 200 & 0.0001 & Adam & 16 & 10 & 200 & 0.01 & Adam & 16 & 10 & 200 \\
        \footnotesize \textsc{ResNet-50 (H)} & 0.01 & Adam & 16 & 10 & 200 & 0.01 & Adam & 16 & 1 & 200 & 0.01 & Adam & 16 & 10 & 200 \\
        \footnotesize \textsc{ViT (H)} & 0.01 & Adam & 16 & 10 & 200 & 0.0001 & Adam & 1 & 10 & 200 & 0.01 & SGD & 1 & 10 & 0 \\
        \midrule
        & \multicolumn{5}{|c|}{\footnotesize \textsc{Continual Online}} & \multicolumn{5}{|c|}{\footnotesize \textsc{Continual Online}} & \multicolumn{5}{|c}{\footnotesize \textsc{Continual Online}} \\
        & \footnotesize $\mathtt{lr}$ & \footnotesize $\mathtt{optim}$ & \footnotesize $b$ & \footnotesize $\mathtt{ep}$ & \footnotesize $|\mathcal{R}|$ & \footnotesize $\mathtt{lr}$ & \footnotesize $\mathtt{optim}$ & \footnotesize $b$ & \footnotesize $\mathtt{ep}$ & \footnotesize $|\mathcal{R}|$ & \footnotesize $\mathtt{lr}$ & \footnotesize $\mathtt{optim}$ & \footnotesize $b$ & \footnotesize $\mathtt{ep}$ & \footnotesize $|\mathcal{R}|$ \\
        \midrule
        \footnotesize \textsc{MLP} & 0.01 & SGD & 1 & 1 & 0 & 0.01 & SGD & 1 & 1 & 200 & 0.0001 & Adam & 16 & 1 & 0 \\
        \footnotesize \textsc{CNN} & 0.0001 & Adam & 1 & 1 & 200 & 0.0001 & Adam & 1 & 1 & 200 & 0.0001 & Adam & 1 & 1 & 200 \\
        \footnotesize \textsc{ResNet-50} & 0.0001 & Adam & 16 & 1 & 200 & 0.0001 & Adam & 16 & 1 & 200 & 0.0001 & Adam & 16 & 1 & 200 \\
        \footnotesize \textsc{ResNet-50 (H)} & 0.01 & Adam & 16 & 1 & 200 & 0.01 & Adam & 16 & 1 & 200 & 0.01 & Adam & 16 & 1 & 200 \\
        \footnotesize \textsc{ViT (H)} & 0.01 & Adam & 16 & 1 & 200 & 0.01 & SGD & 1 & 1 & 0 & 0.01 & Adam & 16 & 1 & 200 \\
        \bottomrule
    \end{tabular}
    \vskip 2mm
    \caption{\textbf{KANDY-Easy} (first column), \textbf{KANDY-Hard} (second column), \textbf{KANDY-Hard-Large} (third column), optimal values of the hyper parameters for neural experiments.}
    \label{tab:hyper}
\end{table}

\parafango{Symbolic models} For symbolic baselines, we performed two separate grid searches for either curriculum. Experiments were repeated with two different engines (Aleph and Popper), on two encodings (natural, \textsc{Nat} and pointer, \textsc{Ptr}) and three background knowledge sizes (minimal, \textsc{Min}, mid-size, \textsc{Mid}, large). Due to different behavior of timeout parameters of the two engines, we impose a hard constraint on time limit by forcefully terminating the engines on timeout.\footnote{This has the side effect of losing any-time partial solutions, but their performance proved to be unsatisfactory in preliminary tests with longer timeouts.}
Unlike experiments with neural models, data for symbolic-model-based experiments was not re-balanced prior to training, moreover, as both Aleph and Popper engines are deterministic, experiments were not repeated multiple times.
\begin{table}[t]
    \centering
    \begin{tabular}{l|l|c}
        \textsc{\small Param} & \textsc{\small Description} & \textsc{\small Value} \\
        \midrule
        max\_literals & Total amount of literals allowed in the theory (Popper) & 40 \\
        min\_acc & Minimum acceptable accuracy (Aleph) & 0.5 \\
        noise & Maximum number of false positives (Aleph) & 5 \\
        i & Layers of new variables (Aleph) & 5 \\
        min\_pos & Minimum number of positives covered by a rule (Aleph) & 2 \\
        depth & Proof depth (Aleph) & 15 \\
        nodes & Nodes to explore during search (Aleph) & 20000 \\
        \bottomrule
    \end{tabular}
    \vskip 3mm
    \caption{Constant hyper-parameters for symbolic experiments.}
    \label{tab:constant_hyper}
\end{table}
Table \ref{tab:constant_hyper} contains hyper-parameters which were kept constant for both curricula.
The Easy curriculum is characterized by a more restrictive set of hyper-parameters (Table \ref{tab:symbolic_hyper} top), as solutions are expected to be significantly simpler than the Hard one (Table \ref{tab:symbolic_hyper} bottom).
Of course, runs with invalid combinations of parameters for a given model (i.e., Aleph with any of singleton\_vars, predicate\_invention or recursion set to True) are not considered.
Instances with Popper either terminated with unsatisfactory accuracy ($< 0.5$ in the training set) or timed out for either curriculum. Likewise, instances using the large KB failed to converge within the allotted time.
Best models were selected by average micro-accuracy on the validation set, grouped by encoding and background knowledge size, after filtering out solutions with accuracy below $0.5$ (i.e., every Popper and cheat background knowledge solution, plus some other combination of hyper-parameters).

\begin{table}[t]
    \centering
    \begin{minipage}{0.02\textwidth}
    \rotatebox{90}{\textbf{Hard} \hskip 1.7cm \textbf{Easy} \hskip 0.3cm}
    \end{minipage}
    \begin{minipage}{0.7\textwidth}
    \begin{tabular}{l|l|c}
        \textsc{\small Param} & \textsc{\small Description} & \textsc{\small Values} \\
        \midrule
        timeout & Time allowed for a single task (in seconds) & 1800 \\
        max\_vars & Maximum number of variables in a clause & 6, 12 \\
        max\_clauses & Maximum number of clauses in the theory & 1 \\
        max\_size & Maximum number of literals in a clause & 6 \\
        singleton\_vars & Enable use of \_ variables (Popper) & True, False  \\
        predicate\_invention & Enable predicate invention (Popper) & True, False \\
        recursion & Enable recursive clauses (Popper) & True, False \\
        \midrule
        timeout & Time allowed for a single task (in seconds) & 3600 \\
        max\_vars & Maximum number of variables in a clause & 12, 15 \\
        max\_clauses & Maximum number of clauses in the theory & 2, 4 \\
        max\_size & Maximum number of literals in a clause & 6, 12, 15 \\
        singleton\_vars & Enable use of \_ variables (Popper) & True, False  \\
        predicate\_invention & Enable predicate invention (Popper) & True, False \\
        recursion & Enable recursive clauses (Popper) & True, False \\
        \bottomrule
    \end{tabular}
    \end{minipage}
    \vskip 3mm
    \caption{Values of hyper-parameters in symbolic models that were evaluated during the cross-validation procedure.}
    \label{tab:symbolic_hyper}
\end{table}

\begin{table}[t]
    \centering
    \begin{tabular}{l|ccc|ccc}
        \multicolumn{1}{c}{$\ $} & \multicolumn{3}{c}{\textbf{Easy}} & \multicolumn{3}{c}{\textbf{Hard}} \\
        \toprule
        & \multicolumn{3}{|c|}{\footnotesize \textsc{Independent}} & \multicolumn{3}{|c}{\footnotesize \textsc{Independent}} \\
        & \footnotesize $\mathtt{vars}$ & \footnotesize $\mathtt{clauses}$ & \footnotesize $\mathtt{size}$ & \footnotesize $\mathtt{vars}$ & \footnotesize $\mathtt{clauses}$ & \footnotesize $\mathtt{size}$  \\
        \midrule
        \footnotesize \textsc{Nat-Min} & 12 & 1 & 6 & 15 & 2 & 15 \\
        \footnotesize \textsc{Nat-Mid} & 12 & 1 & 6 & 12 & 2 & 6 \\
        \footnotesize \textsc{Ptr-Min} & 6 & 1 & 6 & 12 & 2 & 12 \\
        \footnotesize \textsc{Ptr-Mid} & 6 & 1 & 6 & 15 & 4 & 6 \\
        \bottomrule
    \end{tabular}
    \vskip 2mm
    \caption{\textbf{KANDY-Easy} (first column), \textbf{KANDY-Hard} (second column), optimal values of the hyper parameters for symbolic experiments.}
    \label{tab:hyper}
\end{table}

\section{Metrics}
\label{sec:appendix_metrics}
For each task, we measured the accuracy $\mathrm{acc}_j$, which is the micro-accuracy of the binary classification problem of the $j$-th task. It is obtained by averaging the ratio of right predictions for the positive class over the set of positive examples, and the ratio of right predictions for the negative class over the set of negative examples. It can be written as
$$
\mathrm{acc}_j = \frac{1}{2} \cdot \frac{TP_j}{TP_j + FN_j} + \frac{1}{2} \cdot \frac{TN_j}{TN_j + FP_j}
$$
where $TP_j$, $TN_j$, $FP_j$, $FN_j$ is the number of true positives, true negatives, false positives, false negatives for task $j$, respectively.
In continual learning settings, data is streamed in a task-oriented order, thus the data of task $j=0$ first, then the data of task $j=1$ and so on and so forth, up to the last task $j=N-1$. Metrics can be computed at different ``time'' instants, i.e., using different states of the neural models. For example, the model at ``time'' $z$ is the neural model that has learned using data streamed up to task $z$ (included). We use the (overloaded) notation $\mathrm{acc}_j^t$ to indicate the accuracy on task $j$ measure with the model that has learned using data streamed up to task $z$.
Following the definitions in \citep{mai2022online}, the {\sc\small Average Accuracy} at time $z$ is defined as the average of the task-related accuracies exploiting the neural model that was developed up to the $z$-th task, 
$$
\mathrm{average\_accuracy}(z) = \frac{1}{z} \sum_{j=0}^{z} \mathrm{acc}_j^{z},
$$
where $\mathrm{acc}_j^{z}$ is the already introduced micro-accuracy of the binary classification problem of the $j$-th task.
The {\sc\small Average Forgetting} at time $z$ is defined as
$$
\mathrm{average\_forgetting}(z) = \frac{1}{z-1} \sum_{j = 0}^{z-1} \left( \mathrm{acc}_{i}^{\star} - \mathrm{acc}_j^{z} \right),
$$
where $\mathrm{acc}^{\star}_j$ is the best accuracy obtained on task $j$ so far, i.e., $\max_{k \in \{1,\ldots,z-1\}}\mathrm{acc}_j^{k}$.
The (positive) {\sc\small Backward Transfer} at time $z$ measures the (positive) influence on making predictions on those task-data that featured the previously processed tasks,
$$
\mathrm{backward\_transfer}(z) = \max\left( \frac{2}{z(z-1)} \sum_{j=1}^{z} \sum_{h=0}^{j-1}(\mathrm{acc}_j^{h} - \mathrm{acc}_j^j), 0\right).
$$
Finally, {\sc\small Forward Transfer} measures how learning time $z$ (positively) influences predictions in data of future tasks,
$$
\mathrm{forward\_tranfer}(z)= \frac{2}{z(z-1)}  \sum_{j=0}^{z-1} \sum_{h={j+1}}^{z}( \mathrm{acc}_j^{h} - \mathrm{acc}_j^\Delta),
$$
where $\mathrm{acc}_j^{\Delta}$ is the accuracy of a random guess (that we set to $0.5$, for the sake of simplicity, considering that our accuracy measure is class-balanced). Notice that the definition in \citep{mai2022online} is about problems with several mutually exclusive categories and fully labeled data, where the accuracy of a random guess is very low and $\mathrm{acc}_j^\Delta$ is not present at all. This is not our case, where we have task-oriented data with partial labeling, so that we computed the metric exactly as in \citep{mai2022online} and subtracted $\mathrm{acc}_j^{\Delta}=0.5$ afterwards.

\section{Additional details on data generation}
\label{uff}
In this section we provide further details on the way KANDY generates data. A specific and detailed description of each supported operand is provided in the code repository.

\subsection{Sample generation}

Our generator tries to diversify symbolic observations as much as possible, while guaranteeing at the same time that each task in the curriculum meets a target data set size. To achieve both these properties, symbol generation is performed in three phases: ($i.$) rejection sampling, ($ii.$) data set splitting, and ($iii.$) sample rendering.

During the first phase, first a set (positive or negative) is selected with 0.5 probability, then one element from the set is sampled,\footnote{In YAML files sets are specified as top-level lists, thus sampling simply corresponds to randomly selecting a single top-level element.} and instantiated by performing pre-grounding list expansions, grounding and post-grounding list expansions.
At this point, the symbolic representation contains only compositional operators and atomic objects, and it is converted to a string, (which is a 1-to-1 serialization of the symbol) checked against a set of previously generated symbols for that specific task (to avoid redundancies) and, optionally, against a user-defined rule that consists in a disjunction of Horn clauses in the form $\texttt{valid(Sample) :- tail}$ evaluable by a Prolog interpreter (to guarantee task consistency).
A sample is rejected either if it has already been generated (rejection by repetition), or if its sampling set is discordant with the ground truth rule (i.e., a sample from the positive set does not satisfy the rule, or a sample from the negative set satisfies it, rejection by rule). The maximum number of rejections is governed by a patience hyper-parameter.
This sampling process is repeated until the symbol set has a cardinality equal to the target size, or if a sample has exhausted patience (since symbols are sampled uniformly, as soon as a single sample fails for $patience$ consecutive times, it is likely that subsequent samples will fail as well, hence we stop generation altogether).
Training, validation and test sets are built by disjointly splitting the symbol set according to target percentages, this guarantees that no sample leakage occurs between sets.
In the final phase, exactly $\text{target samples} \cdot \text{split percentage}$ are output: if patience was not exhausted in the first phase, sampling is performed without replacement (and symbols are guaranteed to be unique in every split), otherwise, sampling is performed with replacement (and repetitions can only happen inside the same split, with no inter-split leakage).
In the former case, splits are naturally balanced between positive and negative labels (thanks to the 0.5 probability of choosing between positive and negative sets), while in the latter, label distribution will be unbalanced in favor of the label with the highest symbolic variability: if both distributions are exhausted after reaching the target number of samples, imbalance will reflect the ratio of variability between positive and negative sets, if either is still capable of providing new samples, label distribution will be a poor proxy of symbol distribution.

\subsection{Supervision generation}

\begin{figure}[!ht]
    \centering
    \resizebox{0.6\columnwidth}{!}{
    \begin{tikzpicture}[domain=0:1]
             \draw[->] (-0.2,0) -- (5.2,0) node[right] {$t$};
  \draw[->] (0,-0.2) -- (0,3.2) node[above] {\footnotesize $P(supervised|t)$};

  \draw[color=red, dashed]   plot ({\x * 5},{0.7 * exp(-0.7 * \x * (ln(0.7/0.3)/0.7)) * 3});
\draw[color=blue, dashed]   plot (\x * 5, 0.7 * 3) node[above] {\footnotesize $\lambda = 0.7$};
\draw[color=blue, dashed]   plot (\x * 5, 0.3 * 3) node[below] {\footnotesize $\beta = 0.3$};
  
\draw plot[ycomb, mark=*, mark options={color=red, fill=white}] coordinates{(0.0, 2.09999)(0.5, 1.9297)(1.0, 1.77301)(1.5, 1.6289)(2.0, 1.49657)(2.5, 1.37494)(3.0, 1.26343)(3.5,
1.1607)(4.0, 1.06636)(4.5, 0.97966)(5.0, 0.90005)} node[below, color=red] {};

\end{tikzpicture}
}
\caption{An example of supervision schedule with $\gamma=0.7, \beta=0.3$, decayed over 11 samples.}
    \label{suppy}
\end{figure}
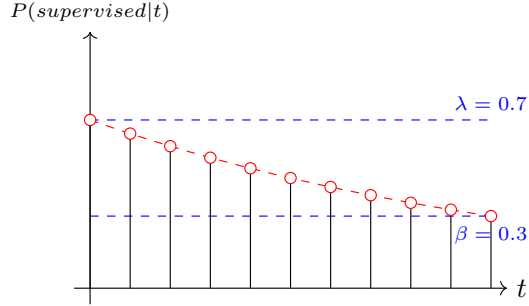

For each task, we assume the learner has at its disposal two objective functions, one to be used when supervisions are available and another when unsupervised discovery is required (unless the user opts for the fully-supervised datasets). We define a task progression variable $t \in [0,1]$, which is instantiated for each (non-rejected) sample in a given task of the curriculum (different tasks have their own private progression variable). We then evaluate the following exponential decay function, evaluated at each step: $f(t) = \gamma \cdot e^{-\sigma \cdot t}$, where $\sigma = \log(\gamma \cdot \beta^{-1})$ and $\gamma, \beta \in [0, 1]$ are hyper-parameters which can be defined independently for each task in the curriculum.
Function $f(t)$ results to be monotonically decreasing and has boundary values $f(0) = \gamma$ and $f(1) = \beta$, see Figure~\ref{suppy} for an example.
For each sample in a given task, we sample a binary random variable with probability $p_t = f(t)$. If the random variable is positive, we mark the sample as supervised, otherwise as unsupervised.
This simple strategy effectively schedules sporadic supervisions in a way which is most beneficial for the learning process: supervisions are more abundant at the beginning, when parameters are far from optimal, and are gradually reduced as learning progresses.
We claim that this strategy offers two main advantages: with only two hyper-parameters it allows to govern the exploration-exploitation trade-off (by assigning different importance to the unsupervised and supervised objectives, respectively, at different time steps), and, in learning settings where annotations are not easily available, it allows to allocate limited annotations in an effective order.

\subsection{Image rendering}
\begin{figure}
    \centering
\resizebox{1.0\columnwidth}{!}{\begin{tikzpicture}[auto, node distance=0.5cm and 2cm,>=latex']
    \draw [thick,dashed] (5.5,1) -- (5.5,-8);
    \node [] (in1) {in(};
    \node [rectangle, right=0.1em of in1, fill=blue, minimum width=2em, minimum height=2em] (s1) {};
    \node [right=0.1em of s1] (in2) {,};
    \node [circle, fill=red, right=0.1em of in2] (c1) {};
    \node [right=0.1em of c1] (in3) {)};
    \node [circle, fill=red, right=4em of in3] (c2) {};
    \begin{pgfonlayer}{background}
    \node [rectangle, minimum width=2em, minimum height=2em, fit=(c2), fill=blue] (s2) {};
    \end{pgfonlayer}

    \node [right=6em of s2] (rnd1) {random(};
    \node [rectangle, right=0.1em of rnd1, fill=blue, minimum width=2em, minimum height=2em] (s3) {};
    \node [right=0.1em of s3] (rnd2) {,};
    \node [circle, fill=red, right=0.1em of rnd2] (c3) {};
    \node [right=0.1em of c3] (rnd3) {)};
    \node [rectangle, minimum width=2em, minimum height=2em, right=4em of rnd3, fill=blue] (s4) {};
    \node [circle, fill=red, above right=-0.4em of s4] (c4) {};

    \node [below=4em of in1] (stk1) {stack(};
    \node [rectangle, right=0.1em of stk1, fill=blue, minimum width=2em, minimum height=2em] (s5) {};
    \node [right=0.1em of s5] (stk2) {,};
    \node [circle, fill=red, right=0.1em of stk2] (c5) {};
    \node [right=0.1em of c5] (stk3) {)};
    \node [rectangle, minimum width=2em, minimum height=2em, below=3em of c2, fill=blue] (s6) {};
    \node [circle, fill=red, below=of s6] (c6) {};

    \draw[draw,->] (s6.center) -- (c6.center);

    \node [below=4em of rnd1] (sbs1) {side\_by\_side(};
    \node [rectangle, right=0.1em of sbs1, fill=blue, minimum width=2em, minimum height=2em] (s7) {};
    \node [right=0.1em of s7] (sbs2) {,};
    \node [circle, fill=red, right=0.1em of sbs2] (c7) {};
    \node [right=0.1em of c7] (sbs3) {)};
    \node [rectangle, minimum width=2em, minimum height=2em, below=4em of s4, fill=blue, xshift=-1.5em] (s8) {};
    \node [circle, fill=red, right=2em of s8] (c8) {};

    \draw[draw,->] (s8.center) -- (c8.center);

    \node [below=4em of stk1] (ullr1) {diag\_ul\_lr(};
    \node [rectangle, right=0.1em of ullr1, fill=blue, minimum width=2em, minimum height=2em] (s9) {};
    \node [right=0.1em of s9] (ullr2) {,};
    \node [circle, fill=red, right=0.1em of ullr2] (c9) {};
    \node [right=0.1em of c9] (ullr3) {)};
    \node [rectangle, minimum width=2em, minimum height=2em, below=3em of s6, fill=blue] (s10) {};
    \node [circle, fill=red, below right=2em of s10] (c10) {};

    \draw[draw,->] (s10.center) -- (c10.center);

    \node [below=4em of sbs1] (urll1) {diag\_ll\_ur(};
    \node [rectangle, right=0.1em of urll1, fill=blue, minimum width=2em, minimum height=2em] (s11) {};
    \node [right=0.1em of s11] (urll2) {,};
    \node [circle, fill=red, right=0.1em of urll2] (c11) {};
    \node [right=0.1em of c11] (urll3) {)};
    \node [circle, fill=red, below=4em of s8, xshift=3em] (c12) {};
    \node [rectangle, minimum width=2em, minimum height=2em, below left=2em of c12, fill=blue] (s12) {};

    \draw[draw,->] (s12.center) -- (c12.center);

    \node [below=4em of ullr1, xshift=-3em] (grd1) {grid(};
    \node [rectangle, right=0.1em of grd1, fill=blue, minimum width=2em, minimum height=2em] (s13) {};
    \node [right=0.1em of s13] (grd2) {,};
    \node [circle, fill=red, right=0.1em of grd2] (c13) {};
    \node [right=0.1em of c13] (grd3) {,};
    \node [regular polygon, regular polygon sides=3, fill=green, right=0.1em of grd3] (t1) {};
    \node [right=0.1em of t1] (grd4) {,};
    \node [rectangle, fill=magenta, right=0.1em of grd4, minimum width=1em, minimum height=1em] (s14) {};
    \node [right=0.1em of s14] (grd5) {)};

    \node [rectangle, right=1.5em of grd5, fill=blue, minimum width=2em, minimum height=2em] (s15) {};
    \node [circle, fill=red, right=1em of s15] (c14) {};
    \node [regular polygon, regular polygon sides=3, fill=green, below=1em of s15] (t2) {};
    \node [rectangle, fill=magenta, right=1.2em of t2, minimum width=1em, minimum height=1em] (s16) {};

    \draw[draw,->] (s15.center) -- (c14.center);
    \draw[draw,->] (c14.center) -- (t2.center);
    \draw[draw,->] (t2.center) -- (s16.center);
    
\end{tikzpicture}}
    \caption{Compositional operator primitives and their graphical rendering. Lists are parsed from left to right and recursively drawn in a bottom up fashion. Not shown: \texttt{quadrant\_\{ul,ur,ll,lr\}} operators behave like \texttt{in}, but offset children in one of the four corners, \texttt{stack\_reduce\_bb} and \texttt{side\_by\_side\_reduce\_bb} behave like their counterparts, but differ on how bounding box sizes are computed).}
    \label{fig:primitives}
\end{figure}
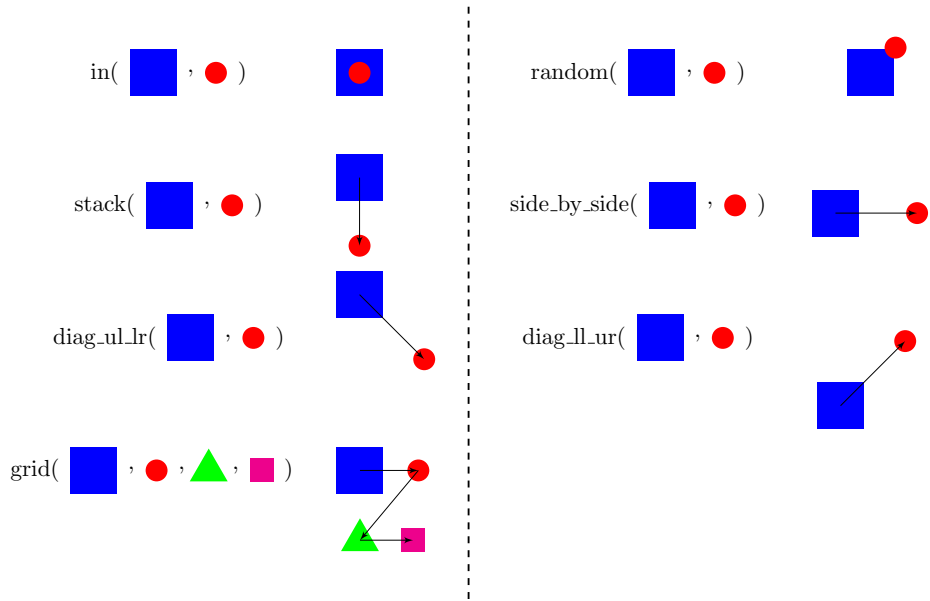

As stated in the main text of the paper, rendering is performed at a constant scale/size, no matter how deep in a hierarchy an atomic shape is.
We deemed alternative approaches not viable, namely, recursively reducing atomic shapes would have caused the risk of shapes becoming unrecognizable at a too-deep level and the inhability of assigning clear labels to different sizes (as a small triangle on the first level of a hierarchy could be mistaken for a large triangle on the second level, and so on), and cropping would have, likewise, caused unrecognizable shapes (with the degenerate case of bounding boxes smaller than the largest inscribed square in a triangle/circle degenerating to a square).
Figure \ref{fig:primitives} shows how compositional primitives render their children.

\subsection{Some examples}
In this section we present four selected examples of task description in the KANDY format, complete with their set generation rule definitions, positive and negative example images, the Prolog ground-truth/rejection rule and some noteworthy ILP solutions.
Set definitions for every task in the curricula are available in the code repository (YAML files), section \ref{sec:appendix_rules} contains the entirety of ground-truth/rejection rules for both curricula (which are also included in YAML files).
Although the generator could produce datasets with just rule set definitions, we create larger, but easier to specify, forests of symbols, which start as supersets of the intended task and are then pruned by rule-based rejection. This ``trick'' is exploited mostly for negative sets. Note, however, that rejections skew distributions towards imbalanced datasets, so this ``overspecify-then-reject'' approach should be used with caution.
To achieve even more compact definitions, we fully exploit YAML anchor, alias and override constructs.

\subsubsection{Easy curriculum - Task 12}
\noindent\textsc{Textual description:}\\
A horizontal line of 3-6 objects such as at least one of them is blue and the right-most is a red triangle.

$\ $\\ \noindent\textsc{Positive set:}
\begin{minted}{yaml}
- side_by_side:
  - permute:
    - {shape: ~, color: blue, size: ~}
    - random_repeat_before:
        min: 1
        max: 4
        list:
          - {shape: ~, color: ~, size: ~}
  - {shape: triangle, color: red, size: ~}
\end{minted}

$\ $\\ \noindent\textsc{Negative set:}
\begin{minted}{yaml}
- side_by_side: # Negative 1: the triangle is not at the end. Note the indentation of 
                # the last line wrt positive set.
    - permute:
        - { shape: ~, color: blue, size: ~ }
        - random_repeat_before:
            min: 1
            max: 4
            list:
              - { shape: ~, color: ~, size: ~ }
        - { shape: triangle, color: red, size: ~ }
- side_by_side: # Negative 2: There is no red triangle but a blue shape
    - permute:
        - { shape: ~, color: blue, size: ~ }
        - pick:
            n: 1
            list:
              - { shape: not_triangle, color: ~, size: ~ }
              - { shape: triangle, color: not_red, size: ~ }
        - random_repeat_before:
            min: 1
            max: 4
            list:
              - { shape: ~, color: ~, size: ~ }
- side_by_side: # Negative 3: There is a red triangle but no blue shape
    - permute:
        - { shape: ~, color: not_blue, size: ~ }
        - random_repeat_before:
            min: 1
            max: 4
            list:
              - { shape: ~, color: ~, size: ~ }
    - { shape: triangle, color: red, size: ~ }
- side_by_side: # Negative 4: There is no blue shape nor red triangle
    - random_repeat_before:
        min: 2
        max: 5
        list:
          - { shape: ~, color: not_blue, size: ~ }
    - pick:
        n: 1
        list:
          - { shape: not_triangle, color: ~, size: ~ }
          - { shape: triangle, color: not_red, size: ~ }
\end{minted}

\vspace{5pt}
$\ $\\ \noindent\textsc{Positive/negative images:}
\begin{figure}[H]
    \centering
    \begin{subfigure}[b]{.45\textwidth}
    \centering
    \includegraphics[width=.45\textwidth]{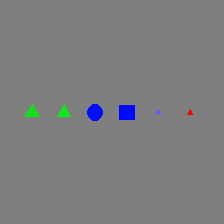}
    \caption{Positive example: the rightmost object is a red triangle and there are three blue objects.}
    \end{subfigure}
    \hfill
    \begin{subfigure}[b]{.45\textwidth}
    \centering
    \includegraphics[width=.45\textwidth]{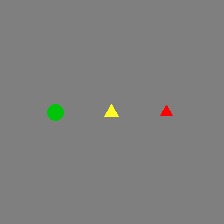}
    \caption{Negative example: the rightmost object is a red triangle, but there are no blue objects.}
    \end{subfigure}
    \caption{Examples for Easy curriculum task 12.}
    \label{fig:task12easy}
\end{figure}

$\ $\\ \noindent\textsc{Ground truth/rejection rule:}
\begin{minted}{prolog}
valid(C) :- extract_children(C, L), 
            last(L, C1), 
            member(C2, L), 
            extract_shape(C1, triangle), 
            extract_color(C1, red), 
            extract_color(C2, blue).
\end{minted}

$\ $\\ \noindent\textsc{Rule inferred by \textsc{Ptr-Min} {\footnotesize(correct but incomplete, precision $0.75$, recall $1.0$)}}
\begin{minted}{prolog}
valid(A) :- sample_is(A,B),
            defined_as(B,side_by_side,C),
            exists_color(red,C),
            exists_color(blue,C).
\end{minted}

\subsubsection{Easy curriculum - Task 13}
\noindent\textsc{Textual description:}\\
Two non-centered objects such as they are a triangle and a square with the same color.

$\ $\\ \noindent\textsc{Positive set:}
\begin{minted}{yaml}
- quadrant_or_center:
  - any_displacement:
    - store_before:
        alias: color
        list:
        - pick_before:
            n: 1
            list:
            - {shape: triangle, color: red, size: ~}
            - {shape: triangle, color: green, size: ~}
            - {shape: triangle, color: blue, size: ~}
            - {shape: triangle, color: cyan, size: ~}
            - {shape: triangle, color: magenta, size: ~}
            - {shape: triangle, color: yellow, size: ~}
    # Convoluted way to extract the color from the stored triangle 
    # and transfer it to a square
    - union: # A union of a small square with the stored color and a large square 
             # with the stored color.
      - difference: # The negation of a non-small non-square with a non-stored color.
        - {shape: ~, color: ~, size: ~}
        - symmetric_difference: # A non-small non-square with a non-stored color.
          - symmetric_difference: # A non-small non-triangle of non-stored color.
            - symmetric_difference: # A non-small negation of the stored object 
                                    # (triangle of specific color).
              - {shape: ~, color: ~, size: small}
              - recall:
                  alias: color
            - {shape: triangle, color: ~, size: ~}
          - {shape: square, color: ~, size: ~}
      - difference:
        - {shape: ~, color: ~, size: ~}
        - symmetric_difference:
          - symmetric_difference:
            - symmetric_difference:
              - {shape: ~, color: ~, size: large}
              - recall:
                  alias: color
            - {shape: triangle, color: ~, size: ~}
          - {shape: square, color: ~, size: ~}
\end{minted}

$\ $\\ \noindent\textsc{Negative set:}
\begin{minted}{yaml}
- quadrant_or_center: # Negative 1: A triangle and a square, but arbitrary colors 
                      # (the rejection rule will remove invalid combinations)
  - any_displacement:
    - {shape: triangle, color: ~, size: ~}
    - {shape: square, color: ~, size: ~}
- quadrant_or_center: # Negative 2: Arbitrary objects
  - any_displacement:
    - { shape: ~, color: ~, size: ~ }
    - { shape: ~, color: ~, size: ~ }
- quadrant_or_center: # Negative 3: Same color, but arbitrary shape
    - any_displacement:
        - repeat_before:
            n: 2
            list:
            - pick_before:
                n: 1
                list:
                  - { shape: ~, color: red, size: ~ }
                  - { shape: ~, color: green, size: ~ }
                  - { shape: ~, color: blue, size: ~ }
                  - { shape: ~, color: cyan, size: ~ }
                  - { shape: ~, color: magenta, size: ~ }
                  - { shape: ~, color: yellow, size: ~ }
\end{minted}

$\ $\\ \noindent\textsc{Positive/negative images:}
\begin{figure}[H]
    \centering
    \begin{subfigure}[b]{.45\textwidth}
    \centering
    \includegraphics[width=.45\textwidth]{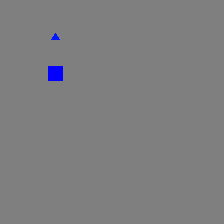}
    \caption{Positive example: there are a blue triangle and a blue square.}
    \end{subfigure}
    \hfill
    \begin{subfigure}[b]{.45\textwidth}
    \centering
    \includegraphics[width=.45\textwidth]{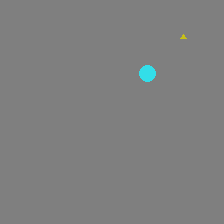}
    \caption{Negative example: there are two circles, violating both the color and shape constraints.}
    \end{subfigure}
    \caption{Examples for Easy curriculum task 13.}
    \label{fig:task13easy}
\end{figure}

$\ $\\ \noindent\textsc{Ground truth/rejection rule:}
\begin{minted}{prolog}
valid(C) :- recursive_contains(C, C1), 
            recursive_contains(C, C2), 
            same_color(_, [C1, C2]), 
            extract_shape(C1, triangle), 
            extract_shape(C2, square).
\end{minted}

$\ $\\ \noindent\textsc{Rule inferred by \textsc{Ptr-Min} {\footnotesize (failure case)}:}
\begin{minted}{prolog}
valid(A) :- sample_is(A,B), 
            recursive_contains(B,C), 
            extract_shape(C,square), 
            extract_size(C,large).
\end{minted}

\subsubsection{Hard curriculum - Task 6}
\noindent\textsc{Textual description:}\\
A collection of compound objects arranged in a diagonal. At least one of them is a ``tower'' (vertical stack of squares with the same size).

$\ $\\ \noindent\textsc{Positive set:}
\begin{minted}{yaml}
- any_diag:
    - permute:
        - repeat_before:
            n: 3
            list: &any_kmer
              - any_non_diag:
                - random_repeat_before:
                    min: 2
                    max: 3
                    list:
                      - { shape: ~, color: ~, size: ~ }
        - pick: &tower
            n: 1
            list:
              - stack:
                  - { shape: square, color: ~, size: small }
                  - { shape: square, color: ~, size: small }
              - stack:
                  - { shape: square, color: ~, size: large }
                  - { shape: square, color: ~, size: large }
              - stack:
                  - { shape: square, color: ~, size: small }
                  - { shape: square, color: ~, size: small }
                  - { shape: square, color: ~, size: small }
              - stack:
                  - { shape: square, color: ~, size: large }
                  - { shape: square, color: ~, size: large }
                  - { shape: square, color: ~, size: large }
\end{minted}

$\ $\\ \noindent\textsc{Negative set:}
\begin{minted}{yaml}
- any_diag:
    - repeat_before:
        n: 4
        list: *any_kmer
\end{minted}

$\ $\\ \noindent\textsc{Positive/negative images:}
\begin{figure}[H]
    \centering
    \begin{subfigure}[b]{.45\textwidth}
    \centering
    \includegraphics[width=.45\textwidth]{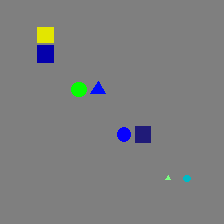}
    \caption{Positive example: there exists a tower in the image (vertical stack of squares).}
    \end{subfigure}
    \hfill
    \begin{subfigure}[b]{.45\textwidth}
    \centering
    \includegraphics[width=.45\textwidth]{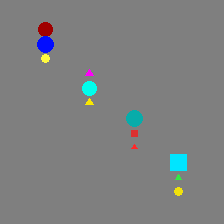}
    \caption{Negative example: there is no tower in the image.}
    \end{subfigure}
    \caption{Examples for Hard curriculum task 6.}
    \label{fig:task6hard}
\end{figure}

$\ $\\ \noindent\textsc{Ground truth/rejection rule:}
\begin{minted}{prolog}
tower(C) :- extract_op_and_chld(C, stack, L),
            same_shape(square, L), 
            same_size(_, L), 
            length(L, N), N >= 2, N =< 3.
valid(C) :- contains(C, C1), tower(C1).
\end{minted}

$\ $\\ \noindent\textsc{Rule inferred by \textsc{Nat-Nrm} {\footnotesize (correct but incomplete)}:}
\begin{minted}{prolog}
valid(A) :-   contains(A,B), 
              extract_op_and_chld(B,stack,C),
              same_shape(square,C).
\end{minted}

\subsubsection{Hard curriculum - Task 15}
\noindent\textsc{Textual description:}\\
A collection of objects such that, if their number is odd, they have the same color, if their number is even, they have the same shape.

$\ $\\ \noindent\textsc{Positive set:}
\begin{minted}{yaml}
- any_displacement: # Odd case
  - permute:
    - store_before:
        alias: color
        list: *fixed_random_color # Defined previously on another task.
    - repeat_before:
        n: 2
        list:
        - random_repeat_before:
            min: 1
            max: 3
            list:
            - recall:
                alias: color
- any_displacement: # Even case
  - permute:
    - repeat_before:
        n: 2
        list:
        - store_before:
            alias: shape
            list: *fixed_random_shape
         - random_repeat_before:
            min: 0
            max: 2
            list:
            - recall:
                alias: shape
\end{minted}

$\ $\\ 
\noindent\textsc{Negative set:}
\begin{minted}{yaml}
- any_displacement: # Odd case
  - permute:
    - store_before:
        alias: shape
        list: *fixed_random_shape
    - repeat_before:
        n: 2
        list:
        - random_repeat_before:
            min: 1
            max: 3
            list:
            - recall:
                alias: shape
- any_displacement: # Even case
  - permute:
    - repeat_before:
        n: 2
        list:
        - store_before:
            alias: color
            list: *fixed_random_color
        - random_repeat_before:
            min: 0
            max: 2
            list:
            - recall:
                alias: color
\end{minted}

$\ $\\ \noindent\textsc{Positive/negative images:}
\begin{figure}[H]
    \centering
    \begin{subfigure}[b]{.45\textwidth}
    \centering
    \includegraphics[width=.45\textwidth]{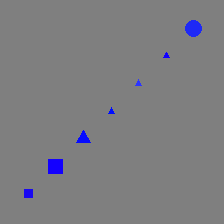}
    \caption{Positive example: there are 7 objects and they share the same color.}
    \end{subfigure}
    \hfill
    \begin{subfigure}[b]{.45\textwidth}
    \centering
    \includegraphics[width=.45\textwidth]{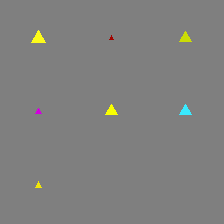}
    \caption{Negative example: there are 7 objects, but they share the same shape.}
    \end{subfigure}
    \caption{Examples for Hard curriculum task 15.}
    \label{fig:task15hard}
\end{figure}

$\ $\\ \noindent\textsc{Ground truth/rejection rule:}
\begin{minted}{prolog}
valid(C) :- extract_children(C, L), length(L, N), odd(N), same_color(_, L).
valid(C) :- extract_children(C, L), length(L, N), even(N), same_shape(_, L).
\end{minted}

$\ $\\ \noindent\textsc{Rule inferred by \textsc{Nat-Nrm} {\footnotesize (failure case)}:}
\begin{minted}{prolog}
valid(A) :- extract_children(A,B), exists_color(green,B), 
            prepend(C,B,D), getmiddle(D,E).
valid(A) :- extract_children(A,B), exists_shape(square,B), 
            exists_shape(circle,B), getmiddle(B,C).
valid(A) :- extract_children(A,B), exists_color(cyan,B),
            prepend(C,B,D), getmiddle(D,E).
valid(A) :- extract_children(A,B), exists_color(yellow,B),
            prepend(C,B,D), dropmiddle(D,E),
            exists_color(red,E).
valid(A) :- extract_children(A,B), dropmiddle(B,C), 
            reverse(C,C), getmiddle(C,D).
valid(A) :- extract_op_and_chld(A,side_by_side,B), 
            getmiddle(B,C), extract_size(C,small).
\end{minted}

\subsection{Background knowledge for ground truth/rejection rules and ILP experiments}
The following is the background knowledge used for rejection sampling during curricula generation. It allows to compactly define every task with at most 6 predicates.
The natural mid-size ({\small\textsc{Nat-Mid}}) background knowledge used in ILP experiments is built from this one, by removing domain-definition predicates (e.g., \texttt{color(red)}) in order to reduce the search space.
$\ $\\

\begin{minted}{prolog}
color(red).
color(green).
color(blue).
color(cyan).
color(magenta).
color(yellow).

shape(triangle).
shape(circle).
shape(square).

size(small).
size(large).

non_diag(stack).
non_diag(side_by_side).
non_diag(stack_reduce_bb).
non_diag(side_by_side_reduce_bb).
non_diag(grid).

diag(diag_ul_lr).
diag(diag_ll_ur).

quadrant_ul(quadrant_ul).
quadrant_ur(quadrant_ur).
quadrant_ll(quadrant_ll).
quadrant_lr(quadrant_lr).

quadrant(quadrant_ul).
quadrant(quadrant_ur).
quadrant(quadrant_ll).
quadrant(quadrant_lr).

quadrant_or_center(in).
quadrant_or_center(X) :- quadrant(X).

non_random(X) :- diag(X).
non_random(X) :- non_diag(X).
non_random(X) :- quadrant(X).
any_composition(X) :- non_random(X).
any_composition(random).


line(X) :- non_diag(X), X \= grid.
line(X) :- diag(X).

house(C) :- extract_op_and_chld(C, stack, [C1, C2]), 
            extract_shape(C1, triangle), 
            extract_shape(C2, square), 
            same_size(_, [C1, C2]).
car(C) :- extract_op_and_chld(C, side_by_side, [C1, C2]), 
          extract_shape(C1, circle), 
          extract_shape(C2, circle), 
          same_size(_, [C1, C2]), 
          same_color(_, [C1, C2]).
tower(C) :- extract_op_and_chld(C, stack, L), 
            same_shape(square, L), 
            same_size(_, L), 
            length(L, N), N >= 2, N =< 3.
wagon(C) :- extract_op_and_chld(C, side_by_side, L), 
            same_shape(square, L), 
            same_size(_, L), 
            length(L, N), N >= 2, N =< 3.
traffic_light(C) :- extract_op_and_chld(C, stack, [C1, C2, C3]), 
                    same_shape(circle, [C1, C2, C3]), 
                    same_size(_, [C1, C2, C3]), 
                    extract_color(C1, red),
                    extract_color(C2, yellow),
                    extract_color(C3, green).
named_object(house).
named_object(car).
named_object(tower).
named_object(wagon).
named_object(traffic_light).
is_named_object(C, house) :- house(C).
is_named_object(C, car) :- car(C).
is_named_object(C, tower) :- tower(C).
is_named_object(C, wagon) :- wagon(C).
is_named_object(C, traffic_light) :- traffic_light(C).

%%%%%%%%%%%%%%%%%%%%%%%%%%%%%%%%%%%%%%%%%%%%%%%%%%%%%%%%%%%%%%%%%%%%
% Natural-mid-size background knowledge starts here:

shape_props(T, SH, CO, SZ) :- atom(T), 
                              term_string(T, S), 
                              split_string(S, "_", "", L),
                              L = [SH, CO, SZ].
extract_shape(T, SH) :- shape_props(T, SH1, _, _), term_string(SH, SH1), shape(SH).
extract_color(T, CO) :- shape_props(T, _, CO1, _), term_string(CO, CO1), color(CO).
extract_size(T, SZ) :- shape_props(T, _, _, SZ1), term_string(SZ, SZ1), size(SZ).

exists_shape(SH, [H|_]) :- extract_shape(H, SH).
exists_shape(SH, [_|T]) :- exists_shape(SH, T).

same_shape(SH, [H]) :- extract_shape(H, SH).
same_shape(SH, [H|T]) :- extract_shape(H, SH), same_shape(SH, T).

exists_color(CO, [H|_]) :- extract_color(H, CO).
exists_color(CO, [_|T]) :- exists_color(CO, T).

same_color(CO, [H]) :- extract_color(H, CO).
same_color(CO, [H|T]) :- extract_color(H, CO), same_color(CO, T).

exists_size(SZ, [H|_]) :- extract_size(H, SZ).
exists_size(SZ, [_|T]) :- exists_size(SZ, T).

same_size(SZ, [H]) :- extract_size(H, SZ).
same_size(SZ, [H|T]) :- extract_size(H, SZ), same_size(SZ, T).

contains(C, X) :- extract_children(C, L), member(X, L).

recursive_contains(C, X) :- contains(C, X), atom(X).
recursive_contains(C, X) :- contains(C, C1), recursive_contains(C1, X).


recursive_contains2(C, X, 0) :- contains(C, X), functor(C, _, 1).
recursive_contains2(C, X, I) :- contains(C, C1), 
                                recursive_contains2(C1, X, J), 
                                I is J + 1.

contains_composition(C, COMP) :- extract_operator(C, COMP).
contains_composition(C, COMP) :- recursive_contains2(C, C1, _),
                                 extract_operator(C1, COMP).

contains_composition_depth(C, COMP, 0) :- extract_operator(C, COMP).
contains_composition_depth(C, COMP, I) :- recursive_contains2(C, C1, J), 
                                          extract_operator(C1, COMP), I is J + 1.

extract_operator(C, COMP) :- functor(C, COMP, 1).
extract_children(C, L) :- functor(C, _, 1), arg(1, C, L).
extract_op_and_chld(C, COMP, L) :- functor(C, COMP, 1), arg(1, C, L).

same_attribute(L) :- same_shape(_, L).
same_attribute(L) :- same_color(_, L).
same_attribute(L) :- same_size(_, L).

same_non_size(L) :- same_shape(_, L).
same_non_size(L) :- same_color(_, L).

all_same(H, [H]).
all_same(H, [H|T]) :- all_same(H, T).


expand2([A, B], A, B).
expand4([A, B, C, D], A, B, C, D).
expand8([A, B, C, D, E, F, G, H], A, B, C, D, E, F, G, H).
expand9([A, B, C, D, E, F, G, H, I], A, B, C, D, E, F, G, H, I).
odd(N) :- N mod 2 =:= 1.
even(N) :- N mod 2 =:= 0.


first([H|_],H).

last([H], H).
last([_|T],X):- last(T, X).


prepend(X, L, [X|L]).
droplast([_], []).
droplast([H|T], [H|T2]):- droplast(T, T2).

middle([_|T], T2):- droplast(T, T2).
getmiddle(L, X) :- length(L, N), odd(N), N1 is div(N, 2), nth0(N1, L, X).
dropmiddle(L, L1) :- getmiddle(L, X), delete(L, X, L1).


less_eq(N, N1) :- N =< N1.
less(N, N1) :- N < N1.
greater(N, N1) :- N > N1.

same(X, Y) :- X = Y.
different(X, Y) :- X \= Y.

% USEFUL BUILT-IN PREDICATES:
% atom(X)
% reverse(L1, L2)
% length(L, N)
% delete(L, X, L1)
% nth0(N, L, X)
% member(X, L)

\end{minted}

\subsubsection{Minimal background knowledge}

The following is the natural minimal ({\small\textsc{Nat-Min}}) background knowledge used in ILP experiments.
$\ $\\

\begin{minted}{prolog}
shape_props(T, SH, CO, SZ) :- atom(T), 
                              term_string(T, S), 
                              split_string(S, "_", "", L), 
                              L = [SH, CO, SZ].
extract_shape(T, SH) :- shape_props(T, SH1, _, _), term_string(SH, SH1), shape(SH).
extract_color(T, CO) :- shape_props(T, _, CO1, _), term_string(CO, CO1), color(CO).
extract_size(T, SZ) :- shape_props(T, _, _, SZ1), term_string(SZ, SZ1), size(SZ).

exists_shape(SH, [H|_]) :- extract_shape(H, SH).
exists_shape(SH, [_|T]) :- exists_shape(SH, T).

same_shape(SH, [H]) :- extract_shape(H, SH).
same_shape(SH, [H|T]) :- extract_shape(H, SH), same_shape(SH, T).

exists_color(CO, [H|_]) :- extract_color(H, CO).
exists_color(CO, [_|T]) :- exists_color(CO, T).

same_color(CO, [H]) :- extract_color(H, CO).
same_color(CO, [H|T]) :- extract_color(H, CO), same_color(CO, T).

exists_size(SZ, [H|_]) :- extract_size(H, SZ).
exists_size(SZ, [_|T]) :- exists_size(SZ, T).

same_size(SZ, [H]) :- extract_size(H, SZ).
same_size(SZ, [H|T]) :- extract_size(H, SZ), same_size(SZ, T).

contains(C, X) :- extract_children(C, L), member(X, L).

recursive_contains(C, X) :- contains(C, X), atom(X).
recursive_contains(C, X) :- contains(C, C1), recursive_contains(C1, X).

extract_operator(C, COMP) :- functor(C, COMP, 1).
extract_children(C, L) :- functor(C, _, 1), arg(1, C, L).
extract_op_and_chld(C, COMP, L) :- functor(C, COMP, 1), arg(1, C, L).


% USEFUL BUILT-IN PREDICATES:
% atom(X)
% reverse(L1, L2)
% length(L, N)

\end{minted}

\subsubsection{Cheat predicates}
The following cheat predicates were used the large background knowledge experiments. As search space is significantly increased, these did not prove to be beneficial for ILP experiments.
$\ $\\

\begin{minted}{prolog}
% CHEAT PREDICATES FOR EASY CURRICULUM:

symmetric_list(L) :- reverse(L, L).

house(C) :- extract_op_and_chld(C, stack, [C1, C2]), extract_shape(C1, triangle), 
            extract_shape(C2, square), same_size(_, [C1, C2]).
car(C) :- extract_op_and_chld(C, side_by_side, [C1, C2]), extract_shape(C1, circle),
          extract_shape(C2, circle), same_size(_, [C1, C2]), 
          same_color(_, [C1, C2]).
tower(C) :- extract_op_and_chld(C, stack, L), same_shape(square, L), 
            same_size(_, L), length(L, N), 
            N >= 2, N =< 3.
wagon(C) :- extract_op_and_chld(C, side_by_side, L), same_shape(square, L), 
            same_size(_, L), length(L, N), 
            N >= 2, N =< 3.
traffic_light(C) :- extract_op_and_chld(C, stack, [C1, C2, C3]), 
                    same_shape(circle, [C1, C2, C3]), 
                    same_size(_, [C1, C2, C3]), extract_color(C1, red), 
                    extract_color(C2, yellow), extract_color(C3, green).
named_object(house).
named_object(car).
named_object(tower).
named_object(wagon).
named_object(traffic_light).
is_named_object(C, house) :- house(C).
is_named_object(C, car) :- car(C).
is_named_object(C, tower) :- tower(C).
is_named_object(C, wagon) :- wagon(C).
is_named_object(C, traffic_light) :- traffic_light(C).

% CHEAT PREDICATES FOR HARD CURRICULUM:
forall_shared_shape(C, SH) :- forall(contains(C, C1), 
                                (contains(C1, C2), extract_shape(C2, SH))).
forall_shared_color(C, CO) :- forall(contains(C, C1), 
                                (contains(C1, C2), extract_color(C2, CO))).
forall_shared_named_obj(C, X) :- forall(contains(C, C1), 
                                (contains(C1, C2), is_named_object(C2, X))).

pseudo_palindrome([]).
pseudo_palindrome([_]).
pseudo_palindrome(L) :- middle(L,M),pseudo_palindrome(M),
                        last(L,A),first(L,B), same_shape(_, [A,B]).
pseudo_palindrome(L) :- middle(L,M),pseudo_palindrome(M),
                        last(L,A), first(L,B), 
                        same_color(_, [A,B]).

pseudo_palindrome2([]).
pseudo_palindrome2([_]).
pseudo_palindrome2(L) :- middle(L,M),pseudo_palindrome2(M),
                         last(L,A),first(L,B), 
                         same_shape(_, [A,B]).
pseudo_palindrome2(L) :- middle(L,M),pseudo_palindrome2(M),
                         last(L,A), first(L,B), 
                         same_color(_, [A,B]).
pseudo_palindrome2(L) :- middle(L,M),pseudo_palindrome2(M),
                         last(L,C1), first(L,C2), 
                         is_named_object(C1, X), is_named_object(C2, X).

\end{minted}

\subsubsection{Knowledge-related differences between pointer and natural encodings}
Background knowledge for pointer ({\sc\small Ptr}) and natural ({\sc\small Nat}) encodings  present only few differences. Most notably, the \texttt{contains/2} predicate is implemented differently:
$\ $\\

\begin{minted}{prolog}
% pointer encoding:
contains(C, X) :- defined_as(C, _, L), member(X, L).

% natural encoding:
contains(C, X) :- extract_children(C, L), member(X, L).

extract_operator(C, COMP) :- functor(C, COMP, 1).
extract_children(C, L) :- functor(C, _, 1), arg(1, C, L).
extract_op_and_chld(C, COMP, L) :- functor(C, COMP, 1), arg(1, C, L).
\end{minted}
$\ $\\

\noindent The other difference concerns sample representation:
$\ $\\
\begin{minted}{prolog}
% generated object (Python's dictionary of lists):
% {'in': 
%   [
%      {'grid': 
%         [
%            {'shape': 'circle', 'color': 'red', 'size': 'small'}, 
%            {'shape': 'square', 'color': 'red', 'size': 'large'},
%            {'shape': 'square', 'color': 'red', 'size': 'small'}
%         ]
%      }
%   ]
% }

% natural encoding (example file):
valid(in([grid([circle_red_small, square_red_large, square_red_small])])).


% pointer encoding (background knowledge):
defined_as(c000000, in, [c000001]).
defined_as(c000001, grid, [circle_red_small, square_red_large, square_red_small]).
sample_is(s00001, c000000).

% pointer encoding (example file):
valid(s00001).
\end{minted}

\subsection{Ground truth/rejection rules for Easy/Hard curricula}\label{sec:appendix_rules}
We report the ground truth Prolog rules that support each tasks in the curricula we released with KANDY.

\subsubsection{Ground truth/rejection rules - Easy}
The following are the ground truth/rejection rules used to filter samples during generation of the Easy curriculum.
$\ $\\

\begin{minted}{prolog}
% task 0:
valid(C) :- contains(C, C1), extract_shape(C1, triangle).

% task 1:
valid(C) :- contains(C, C1), extract_shape(C1, square).

% task 2:
valid(C) :- contains(C, C1), extract_shape(C1, circle).

% task 3:
valid(C) :- contains(C, C1), extract_color(C1, red).

% task 4:
valid(C) :- contains(C, C1), extract_color(C1, green).

% task 5:
valid(C) :- contains(C, C1), extract_color(C1, blue).

% task 6:
valid(C) :- contains(C, C1), extract_color(C1, cyan).

% task 7:
valid(C) :- contains(C, C1), extract_color(C1, magenta).

% task 8:
valid(C) :- contains(C, C1), extract_color(C1, yellow).

% task 9:
valid(C) :- extract_children(C, L), last(L, C1), 
            extract_shape(C1, triangle), extract_color(C1, red).

% task 10:
valid(C) :- extract_children(C, L), last(L, C1), 
            extract_shape(C1, triangle), extract_color(C1, red).

% task 11:
valid(C) :- extract_children(C, L), last(L, C1), 
            member(C2, L), extract_shape(C1, triangle), 
            extract_color(C1, red), extract_shape(C2, circle).

% task 12:
valid(C) :- extract_children(C, L), last(L, C1), 
            member(C2, L), extract_shape(C1, triangle), 
            extract_color(C1, red), extract_color(C2, blue).

% task 13:
valid(C) :- recursive_contains(C, C1), recursive_contains(C, C2), 
            same_color(_, [C1, C2]), extract_shape(C1, triangle), 
            extract_shape(C2, square).

% task 14:
valid(C) :- extract_children(C, L), reverse(L, L).

% task 15:
valid(C) :- contains(C, C1), house(C1).

% task 16:
valid(C) :- contains(C, C1), car(C1).

% task 17:
valid(C) :- contains(C, C1), tower(C1).

% task 18:
valid(C) :- contains(C, C1), wagon(C1).

% task 19:
valid(C) :- contains(C, C1), traffic_light(C1).

\end{minted}

\subsubsection{Ground truth/rejection rules - Hard}
The following are the rejection rules used to filter samples during generation of the Hard curriculum. The predicate \texttt{valid/1} is the one used for rejection sampling, other predicates are auxiliary definitions used for shorter notations. Tasks 15 and 16 are characterized by disjunctive clauses of the target predicate, other tasks may require disjunctive clauses for intermediate predicates.
Tasks using the \texttt{forall/2} metapredicate involve first order logic like the others, but may not be described by a set of Horn clauses (traditional ILP methods are thus expected to fail on these tasks).
$\ $\\

\begin{minted}{prolog}
% task 0:
valid(C) :- contains(C, C1), extract_children(C1, L), length(L, 2), same_color(_, L).

% task 1:
valid(C) :- contains(C, C1), extract_children(C1, L), length(L, 2), same_shape(_, L).

% task 2:
valid(C) :- contains(C, C1), extract_children(C1, L), length(L, 3), same_color(_, L).

% task 3:
valid(C) :- contains(C, C1), extract_children(C1, L), length(L, 3), same_shape(_, L).

% task 4:
valid(C) :- contains(C, C1), house(C1).

% task 5:
valid(C) :- contains(C, C1), car(C1).

% task 6:
valid(C) :- contains(C, C1), tower(C1).

% task 7:
valid(C) :- contains(C, C1), wagon(C1).

% task 8:
valid(C) :- contains(C, C1), traffic_light(C1).

% task 9:
valid(C) :- shape(SH), 
            forall(contains(C, C1), 
                (contains(C1, C2), extract_shape(C2, SH))).

% task 10:
valid(C) :- color(CO), 
            forall(contains(C, C1), 
                (contains(C1, C2), extract_color(C2, CO))).

% task 11:
valid(C) :- named_object(X), 
            forall(contains(C, C1), 
                (contains(C1, C2), is_named_object(C2, X))).

% task 12:
valid(C) :- extract_children(C, L), reverse(L, L).

% task 13:
pseudo_palindrome([]).
pseudo_palindrome([_]).
pseudo_palindrome(L) :- middle(L,M), pseudo_palindrome(M),
                        last(L,A), first(L,B), 
                        same_shape(_, [A,B]).
pseudo_palindrome(L) :- middle(L,M), pseudo_palindrome(M), 
                        last(L,A), first(L,B), 
                        same_color(_, [A,B]).
valid(C) :- extract_children(C, L), pseudo_palindrome(L).

% task 14:
pseudo_palindrome2([]).
pseudo_palindrome2([_]).
pseudo_palindrome2(L) :- middle(L,M), pseudo_palindrome2(M),
                         last(L,A), first(L,B), 
                         same_shape(_, [A,B]).
pseudo_palindrome2(L) :- middle(L,M), pseudo_palindrome2(M),
                         last(L,A), first(L,B), 
                         same_color(_, [A,B]).
pseudo_palindrome2(L) :- middle(L,M), pseudo_palindrome2(M),
                         last(L,C1), first(L,C2), 
                         is_named_object(C1, X), is_named_object(C2, X).
valid(C) :- extract_children(C, L), pseudo_palindrome2(L).

% task 15:
valid(C) :- extract_children(C, L), length(L, N), odd(N), same_color(_, L).
valid(C) :- extract_children(C, L), length(L, N), even(N), same_shape(_, L).

% task 16:
tmp(C) :- contains(C, C1), is_named_object(C1, traffic_light).
tmp2(C) :- contains(C, C2), is_named_object(C2, house).

valid(C) :- contains(C, C1), is_named_object(C1, traffic_light), 
            contains(C, C2), is_named_object(C2, car).
valid(C) :- contains(C, C1), is_named_object(C1, house), 
            contains(C, C2), is_named_object(C2, tower).
valid(C) :- not(tmp(C)), not(tmp2(C)).

% task 17:
tmp(C) :- contains(C, C1), is_named_object(C1, traffic_light).
tmp2(C) :- contains(C, C2), is_named_object(C2, house).

valid1(C) :- contains(C, C1), is_named_object(C1, traffic_light), 
             contains(C, C2), is_named_object(C2, car).
valid1(C) :- contains(C, C1), is_named_object(C1, house), 
             contains(C, C2), is_named_object(C2, tower).
valid1(C) :- not(tmp(C)), not(tmp2(C)).

valid(C) :- forall(contains(C, C1), valid1(C1)).
\end{minted}

\bibliography{biblio}